\documentclass{article}

\usepackage{arxiv}

\usepackage[utf8]{inputenc} 
\usepackage[T1]{fontenc}    
\usepackage{hyperref}       
\usepackage{url}            
\usepackage{booktabs}       
\usepackage{amsfonts}       
\usepackage{nicefrac}       
\usepackage{microtype}      
\usepackage{lipsum}

\usepackage{amsmath}
\usepackage{diagbox}
\usepackage{color}
\usepackage{bm}
\usepackage{array}
\usepackage[utf8]{inputenc}
\usepackage{graphicx}
\usepackage{float}
\usepackage{gensymb}
\usepackage{amsmath}
\usepackage{caption}
\usepackage{subcaption}
\usepackage{multirow}

\usepackage{gensymb}
\usepackage{color}
\usepackage{booktabs}
\usepackage[ruled,vlined]{algorithm2e}
\usepackage{makecell}
\newcolumntype{I}{!{\vrule width 1.5pt}}
\usepackage{booktabs}
\newcommand*\diff{\mathop{}\!\mathrm{d}}
\usepackage{graphicx}  
\usepackage{soul}
\usepackage{subcaption}
\usepackage{mwe}

\usepackage{tikz}
\definecolor{fc}{HTML}{1E90FF}
\tikzset{fc/.style={black,draw=black,fill=fc,rectangle,minimum height=1cm}}
\definecolor{h}{HTML}{228B22}
\definecolor{bias}{HTML}{87CEFA}
\tikzset{h/.style={black,draw=black,fill=h,rectangle,minimum height=1cm}}
\tikzset{bias/.style={black,draw=black,fill=bias,rectangle,minimum height=1cm}}
\definecolor{anti-flashwhite}{rgb}{0.95, 0.95, 0.96}
\definecolor{almond}{rgb}{0.98, 0.91, 0.71}

\title{PI-VAE: Physics-Informed Variational Auto-Encoder for stochastic differential equations}

\author{
  Weiheng Zhong \\
  Department of Civil and Environmental Engineering\\
  University of Illinois at Urbana-Champaign\\
  Champaign, Illinois\\
  \texttt{weiheng4@illinois.edu} \\
   \And
 Hadi Meidani \\
  Department of Civil and Environmental Engineering\\
  University of Illinois at Urbana-Champaign\\
  Champaign, Illinois \\
  \texttt{meidani@illinois.edu} \\
}
 
\begin{document}
\maketitle

\begin{abstract}
We propose a new class of physics-informed neural networks, called physics-informed Variational Autoencoder (PI-VAE), to solve stochastic differential equations (SDEs) or  inverse problems involving SDEs. In these problems the governing equations are known but only a limited number of  measurements of  system parameters  are available. PI-VAE consists of a variational autoencoder (VAE), which generates samples of system variables and parameters. This generative model is integrated with the governing equations. In this integration,  the derivatives of VAE outputs are readily calculated using  automatic differentiation, and used in the physics-based loss term. In this work, the loss function is chosen  to be the Maximum Mean Discrepancy (MMD) for improved  performance, and neural network parameters  are updated iteratively using the stochastic gradient descent algorithm. We first test the proposed method on approximating stochastic processes. Then we study three types of problems related to SDEs:  forward and inverse problems together with mixed problems where  system parameters and solutions are simultaneously calculated.  The satisfactory accuracy and efficiency of the proposed method are numerically demonstrated in comparison with  physics-informed generative adversarial network (PI-WGAN).
\end{abstract}


\section{Introduction}
Stochastic differential equations (SDEs) are differential equations that are parameterized by  random variables and/or constrained by uncertain initial and boundary conditions. SDEs are used in various fields, such as mechanics, economics, biology, and physics \cite{braumann2019introduction}. Solving SDEs can be regarded as quantifying the effects that  random inputs have on the system response or quantities of interest (QoI). Because of the stochastic nature of these problems,  time-efficient numerical methods are needed.  The more  traditional solution approaches are Monte Carlo Simulation (MCS) and Generalized Polynomial Chaos (gPC) methods. Monte Carlo simulation (MCS) is straightforward and robust, but comes with  high computational cost. In MCS, we sample realizations from random inputs, and for each input realization a deterministic differential equations is solved using numerical methods such as finite element, finite difference, and finite volume \cite{fishman2013monte}. The statistics of  QoIs can then be calculated using the obtained deterministic solutions. As another solution approach, gPC methods consider a finite-term polynomial representation for the response, which can be calculated using  collocation or Galerkin method \cite{ghanem2003stochastic, xiu2003modeling, elman2011assessment, luthen2021sparse, alemazkoor2017divide}. However, gPC methods, especially in collocation-based form, suffer from the "curse of dimensionality".

Recently, models based on neural networks   have received increasing attention as an effective approach to solve SDEs. In particular, physics-constrained, or physics-informed neural networks, have been widely studied. The idea of using physics-informed training for neural network solutions of differential equations  was first introduced in 1990s, where neural networks solutions for initial/boundary value problems were developed in \cite{dissanayake1994neural,lagaris1998artificial,psichogios1992hybrid}. But it was only recently that it gained much attention, mainly due to advances in computing hardware and libraries, as an efficient solution for deterministic differential equations~\cite{raissi2017physics,weinan2017deep, rudy2017data, rudy2019data}. More recently, ideas based on physics-informed neural networks have also been proposed for solving SDEs. In particular, a physics-informed deep residual network was used to solve random PDEs using physics-informed deep residual networks by constraining the neural network to satisfy governing equations in a strong or variational form \cite{nabian2019deep}. A probabilistic formulation in the framework of Generative adversarial networks (GANs) to perform UQ of SDEs was presented in \cite{yang2019adversarial}. Also, physics laws have been incorporated into Deep operator networks (DeepONets) to learn the solution operator of parametric differential equations, which can predict the solution of a differential equation for any realization of the uncertain parameters \cite{wang2021learning}. These methods can effectively  solve SDEs when the governing differential equations and its parameters are precisely known. 

However, in many practical  problems, we do not have access to  exact analytical representations for  random parameters of SDEs. A more common case is that we know precisely about the governing differential equations, but  can only obtain information about the parameters using a limited number of sensors that capture scattered measurements (e.g., limited samples of  soil permeability obtained from a subsurface region). To address this, Zhang et al. \cite{zhang2019quantifying}  proposed to solve SDEs by training physics-informed neural networks (PINNs) to learn the modal functions of the arbitrary polynomial chaos (aPC) expansion of the solution of the differential equation. In this paper, all the training data are collected by sparsely distributed fixed sensors. Inspired by \cite{zhang2019quantifying},  a Physics-informed Wasserstein Generative Adversarial Networks (PI-WGANs)  was proposed for solving SDEs \cite{yang2018physics} and SPDEs \cite{yang2019highly}, where  sampled data from the parameters were used to train generators in Generative Adversarial Networks (GANs) which in turn produce samples of QoIs, with similar statistics.

Encoding physics laws, in the form of differential equations, into the framework of GANs successfully mitigates the "curse of dimensionality" and provides an accurate estimation. Nevertheless, GANs suffer from training instability because of simultaneous training of generator and discriminator, which sometimes leads to the problem of mode collapse \cite{kushwaha2020study}. Besides, as mentioned in \cite{yang2018physics}, the computational cost of training GANs is much higher than training a feed-forward neural network. Our goal is then to seek a more stable and more efficient physics-informed learning method to solve SDEs. Similar to GANs, Variational Autoencoder (VAE) is another data generative model initially proposed in \cite{kingma2013auto}, and has proved effective for various tasks of reproducing synthetic data \cite{blaauw2016modeling, hennig2017classifying, yamshchikov2020music}.  In this work, we seek to encode physics laws into VAE and  propose Physics-Informed VAE (PI-VAE), which is   a framework for solving SDEs. PI-VAE is easier for training because  VAE has a simple end-to-end structure, similar to feed-forward neural networks. 

 The architecture of  PI-VAE consists of an encoder structure and a physics-informed decoder structure, constructing a bi-directional mapping between latent space and high-dimensional input space. As the loss function, we use the Maximum Mean Discrepancy of the reconstructed data and training data in latent and high-dimensional input spaces over the fixed positions of the entire computational domain. The parameters of the encoder and the decoder are updated by gradient descent steps toward minimizing the loss function, using variants of the mini-batch gradient descent algorithm. Then the solution to the SDE is represented by the well-trained decoder.
  
The remainder of this paper is organized as follows. In Section~\ref{sec.bg}, we  briefly introduce our data-driven problems that are solved in this paper. Section~\ref{sec.methodology} briefly reviews distance measures between probability distributions and their applications in machine learning, and then introduces the main PI-VAE algorithms for approximating random processes and solving SDEs. Finally, a detailed analysis of performance evaluation of the proposed methods and conclusions are included in Sections~\ref{sec.results} and \ref{sec.conclusions}.



\section{Technical background}\label{sec.bg}

Let us consider the following SDE
\begin{align}
\begin{split}
    \mathcal{N}_x [u(x,\omega), k(x,\omega)] &= f(x,\omega) \quad x \in \bm{D}, \quad \omega \in \bm{\Omega}, \\
    \mathcal{B}_x [u(x,\omega)] &= b(x,\omega), \quad x \in \Gamma,
    \end{split}
\end{align}
where $\mathcal{N}_x$ is a general differential operator, $\bm{D}$ is a $d$-dimensional physical domain in $R^d$, $x$  is a d-dimensional spatial coordinate, $\Omega$ is a probability space, and $\omega$ is a random event, and $\mathcal{B}_x$ is a boundary condition operator acting on the domain boundary $\Gamma$. The coefficient $k(x,\omega)$, the forcing term $f(x,\omega)$  and the boundary conditions $b(x,\omega)$  can be considered to be random processes, and thus the solution $u(x,\omega)$ will also be a random process. 

Depending of the availability of measurement data  from $k(x,\omega)$, $u(x,\omega)$, $f(x,\omega)$ and $b(x,\omega)$ we can two different problems, namely  a forward problem which concerns the calculation of  the solution to the SDE $u (x,  \omega)$ as the QoI given measurements of $k(x,\omega)$, $f(x,\omega)$ and $b(x,\omega)$, or an inverse problem which involves estimating  the coefficient  $k (x,  \omega)$ as the QoI given  available measurements of $u(x,\omega)$, $f(x,\omega)$ and $b(x,\omega)$. Also, one can be dealing with ``mixed" problems where only partial knowledge of $k(x,\omega)$ and $u(x,\omega)$ is available. This class includes a spectrum of problems  depending on the number of sensors on $k(x,  \omega)$ versus $u(x,  \omega)$, where by decreasing the number of sensors on $k(x,  \omega)$ and increasing the number of sensors on $u(x,  \omega)$, the estimation problem gradually transforms from forward to a mixed and finally to an inverse problem  \cite{zhang2019quantifying, yang2018physics}. In this work, we also apply our proposed method to high-dimensional stochastic problems, where the uncertainty in the complex system properties or forcing terms can only be characterized by a high-dimensional stochastic space. 

Our approach is based on the VAE framework which involves comparison between probability distributions. There are various distance measures that can be adopted for calculating the distance between two distributions. Our approach uses one of these  measures, namely Maximum Mean Discrepancy (MMD). For comparison, we have also used other measures for probability distributions  in the examples of Section~\ref{sec.results}. In this section, we briefly explain the used measures, which include a kernel-based  and two Optimal Transport  measures. We  also introduce a brief background on VAEs and its improved variants that are proposed in the literature for better performance.

\subsection{Measurement of distance between probability distributions}

Optimal Transport (OT) theory was initially proposed  to determine the optimal way to redistribute a mass to a new location, and was later used widely as a distance measure between distributions \cite{villani2003topics}. Specifically, the distance between two distributions is considered to be the effort of transporting  a mass  distributed according to a function $P_X$  to another location with a different distribution $P_Y$. Kantorovich's formulation of this problem is given by:
\[
  W_c (P_X, P_Y) := \inf_{ \tau \in P(x \sim P_X, x \sim P_Y)} \mathbb{E}_{(X,Y) \sim \tau} [c(x,x)],
\]
where $c(X,Y)$ is any self-defined cost function to measure the cost of moving density from position of distribution $P_X$ to the position of another distribution $P_G$ and $\tau$ is one of the joint distributions of $(X,Y)$ whose marginals are $P_X$ and $P_G$, respectively. Wasserstein distance is a special case of $W_c$ where $c(X,Y)=d^p(X,Y)$ for $p \geq 1$, where $d(\cdot,\cdot)$ is a distance measure defined for realizations of variables $X$ and $Y$. We call $W_p$ as the p-Wasserstein distance, which is the $p$-th root of $W_c$. Particularly when $p=1$, the 1-Wasserstein distance can be simplified as:
\[
  W_1 (P_X, P_Y) := \sup_{f \in F_L}  \mathbb{E}_{X \sim P_X} [f(X)] -  \mathbb{E}_{Y \sim P_Y} [f(Y)]
\]
where $F_L$ is a class of bounded 1-Lipschitz functions on the metric space $(X,d)$. 

Calculating Wasserstein distance is computationally demanding.  As an algorithm to approximate the Wasserstein distance, the Sinkhorn Iteration  was proposed, where the entropy regularized OT cost was calculated  using discrete samples from the distribution. The resulting Sinkhorn divergence   is a biased estimate of the Wasserstein distance \cite{cuturi2013sinkhorn}. It has been shown that Sinkhorn divergence can approximate the Wasserstein distance with satisfactory accuracy at a much smaller computational cost when appropriate hyper-parameters are selected \cite{benamou2015iterative}.

 Maximum Mean Discrepancy (MMD) has also been introduced to measure the distance between probability distributions \cite{gretton2012kernel}. In a data space $\bm{\chi}$, given independent random variables $x, x' \in \bm{\chi}$ with distribution $P$ and independent random variables $y, y' \in \bm{\chi}$ with distribution $Q$, the MMD between $P$ and $Q$ over $\bm{\chi}$ is defined as:
\[
  \text{MMD}^2_k(P,Q) := \mathbb{E}_{x,x'}[k(x,x')] + \mathbb{E}_{y,y'}[k(y,y')] - 2 \mathbb{E}_{x,y}[k(x,y)],
\]
where $k$ is the kernel of a reproducing kernel Hilbert space (RKHS) $H_k$ of functions on $\bm{\chi}$. We note that  $\text{MMD}_k(P,Q)=0$ if and only if $P=Q$.  When MMD  is used as the loss function in stochastic gradient descent-based trainings, an unbiased U-statistic empirical estimator of MMD is calculated using discrete samples.  This estimator for $m$ realizations of distribution $P$ denoted by $\{x^{i}\}_{i=1}^m$ and $n$ realizations of distribution $Q$, denoted by $\{y^{i}\}_{i=1}^n$, using kernel $k$ is given by:
\[
  \text{MMD}^2_{k,U}(\{x^{i}\}_{i=1}^m, \{y^{i}\}_{i=1}^n) := \frac{1}{m(m-1)} \sum_{i=1}^m \sum_{j \neq i}^m k(x^{(i)}, x^{(j)}) + \frac{1}{n(n-1)} \sum_{i=1}^n \sum_{j \neq i}^n k(y^{(i)}, y^{(j)}) - \frac{2}{mn} \sum_{i=1}^m \sum_{j=1}^n k(x^{(i)}, y^{(j)}).
\]

Since MMD is a kernel-based measurement, different choices of characteristic kernels will lead to different test powers for finite sample sizes \cite{sutherland2016generative}. But MMD is a reliable metric, its evaluations with various characteristic kernels have been shown to be consistent \cite{gretton2012kernel}. 

\subsection{Variational autoencoders}

Variational Autoencoders are neural network models that are used for efficient representation of complex data structures, e.g. for low-dimensional representation of high-dimensional data. The AE architecture consists of an encoder that maps the data into a latent space and a decoder that maps points from the latent space back into the original space.  Specifically, let $\mathcal{E} : X \rightarrow Z$ denote a parametric mapping from the input data space $X$ to a latent space $Z$ (e.g., a neural network encoder). Utilizing a technique often used in the theoretical physics community, known as Random Variable Transformation (RVT), the probability density function of the encoded samples $z$ can be expressed in terms of $\mathcal{E}$ and $p_X$ by:
\[
  P_Z(z) = \int_X p_X(x) \delta(z-\mathcal{E}(x)) \diff x 
\]

where $\delta$ denotes the Dirac distribution function. The main objective of the encoder is to encode the input data $x \in X$ into latent variable $z \in Z$ such that x can be recovered from the probability density function of the encoded samples, $p_Z$, following a prior distribution $q_Z$. Similar to traditional autoencoders, a decoder $D: Z \rightarrow X$ is required to map the latent codes back to the input data space such that:
\[
  P_Y(y) = \int_X p_X(x) \delta(y-\mathcal{D}(\mathcal{E}(x))) \diff x
\]

where y denotes the decoded samples. It is straightforward to see that when $\mathcal{E}=\mathcal{D}^{-1}$, the distribution of the decoder $p_Y$ and the input distribution $p_X$ are identical. Hence, the training of an autoencoder involves learning parameters of $\mathcal{E}$ and $\mathcal{D}$ by minimizing a loss functions which consists of dissimilarity measures between $p_Y$ and $p_X$ in input data space and between $p_Z$ and $q_Z$ in latent space. Defining and implementing the dissimilarity measure is an important design decision. 

For comparison in the latent space, traditional VAE uses KL divergence as the dissimilarity measure. However, KL divergence only measures the information loss of approximating one distribution with the other one but can not measure the distance from a distribution to another distribution. Various alternatives exist to address this limitation. In Wasserstein autoencoder \cite{tolstikhin2017wasserstein}, the authors proposed a GAN-based method and maximum mean discrepancy (MMD)-based approach to measure the disparity of latent variables. In the Sinkhorn autoencoder \cite{patrini2020sinkhorn}, the Sinkhorn iteration was used to approximate the Wasserstein distance as the criterion to constrain the distribution of latent variables. 

For comparison in the input data space, various  measures can be used. In traditional variational autoencoders,  the cross-entropy loss or mean square error (MSE) is used in the loss function \cite{kingma2014autoencoding}. The Wasserstein GAN \cite{arjovsky2017wasserstein} used the OT-based  1-Wasserstein distance to measure data similarity in the input data space and reformulated the loss function as a min-max optimization problem solved by adversarial training scheme. MMD is also used in \cite{li2017mmd}, which replaces the discriminator of GAN with a two-sample test based on MMD. Even though there exists limited literature about autoencoders exploring other algorithms for constraining the distribution in input data space, inspired by the above two papers, replacing the current popular loss function with MMD in input data space may improve the performance of autoencoders.

\section{Methodology}\label{sec.methodology}

In this section, we propose two PI-VAE algorithms; one  for approximating a stochastic process and one for solving an SDE. Although approximation of stochastic processes is not the primary objective of PI-VAE, numerical experiments on reconstructing the sample paths of a stochastic process can serve as preliminary tests of PI-VAE on stochastic processes. Following that, we discuss how a PI-VAE  can be constructed and trained to solve forward problems, inverse problems, and mixed problems involving SDEs.

\subsection{Approximating stochastic processes with VAE} \label{sec.method_GP}

We first consider modeling a stochastic process $f(x, \omega)$ on the domain $\bm{D} \in R^d$ using a limited number of measurements from scattered sensors on this process. Specifically, we place $n_f$ sensors at locations $\{x_i\}_{i=1}^{n_f}$ to collect "snapshots" of $f(x,\omega)$. Here, a "snapshot" represents simultaneous record of all sensors, so data in one snapshot correspond to same random event $\omega$. We collect a total of $N$ snapshots of $f(x,  \omega)$, each taken at one realization of $\omega$ and at $n_f$ different sensor locations:
\[
  \{F(\omega^{(j)})\}_{j=1}^N = \{(f(x_i,\omega^{(j)}))_{i=1}^{n_f}\}_{j=1}^N
\]

The VAE framework consists of an encoder $\mathcal{E}_{\phi}(\cdot)$, which  is a feed-forward DNN parameterized by $\phi$ and a decoder $\tilde{f}_{\theta}(\cdot)$, which is a feed-forward DNN parameterized by $\theta$. The encoder takes sensor measurements $F(\omega^{(j)})$ as input and returns the corresponding latent variable $z_j=\mathcal{E}_{\phi}(F(\omega^{(j)}))$ as outputs. Then latent variables $\{z_j\}$ and coordinates $\{x_i\}$ will be then plugged into the decoder as inputs to calculate the reconstructions $\tilde{f}_{\theta}(x_i,z^{(j)})$ as the final outputs. The encoder and decoder will be trained simultaneously in this end-to-end structure with variants of mini-batch gradient descent algorithms based on the following loss function:
\[
  L(\phi, \theta) = \sum_{i=1}^M \{\text{MMD}_{k_i,U}(z,\zeta) + \text{MMD}_{k_i,U}(f(x,\omega),\tilde{f}_\theta(x,z))\}.
\]

The first term in this loss function represents the empirical MMD estimator between the sample distribution of latent variables $z$ and that of random realizations $\zeta$ drawn from the prior distribution, taken to be a normal distribution. The second term is the empirical MMD estimator between sensor measurements and reconstructed data. After the training, the output of a well-trained decoder, $\tilde{f}_{\theta}(x,\zeta)$, can be used to approximate the stochastic process $f(x,\omega)$. Both of the MMD estimators in this loss functions are evaluated at $M$ different Gaussian kernels. The detailed algorithm is presented in Algorithm \ref{alg.sde}.

\begin{algorithm}
\SetAlgoLined
Initialization: Set the number of training steps $n_t$, batch size $N$, Adam hyper-parameters $\alpha, \beta_1, \beta_2$, initial parameters for encoder and decoder $\phi$ and $\theta$, kernels of MMD estimator $\{k_i\}_{i=1}^M$. \\
 \For {$i = 1,2,...,n_t$}
 {
  Sample $N$ snapshots $\{F(\omega^{(j)})\}_{j=1}^N$. \\
  Sample $N$ random vector from Prior $\{\zeta_j\}_{j=1}^N$ . \\
  \For{$j = 1,2,...,n$}
    {
      $z_j = \mathcal{E}_{\phi}(F(\omega^{(j)}))$ \\
      $\tilde{F}(\omega^{(j)}) = \tilde{f}_{\theta}(z_j)$ \\
    }
  $L = \sum_i^M \text{MMD}_{k_i,U}(\{z_j\}_{j=1}^N, \{\zeta_j\}_{j=1}^N) + \text{MMD}_{k_i,U}(\{F(\omega^{(j)})\}_{j=1}^N, \{\tilde{F}(\omega^{(j)})\}_{j=1}^N)$ \\
  $\theta, \phi \leftarrow \text{Adam}(\nabla L, \theta, \phi, \alpha, \beta_1, \beta_2 )$
 }
 \caption{VAE for approximating stochastic process}
 \label{alg.sde}
\end{algorithm}

\subsection{Solving SDEs with PI-VAE}

Let us consider the following SDE
\begin{align}\label{eq.sde}
    \mathcal{N}_x [u(x,\omega), k(x,\omega)] &= f(x,\omega) \quad x \in \bm{D}, \quad \omega \in \bm{\Omega}, \\
    \mathcal{B}_x [u(x,\omega)] &= b(x,\omega), \quad x \in \Gamma,
\end{align}
defined in Section~\ref{sec.bg}. The objective is to calculate the solution $u(x,\omega)$ using measurements from scattered sensors which form snapshots of the three stochastic processes, $k(x,\omega)$, $f(x,\omega)$ and $b(x,\omega)$, at locations $\{ x^k_i\}_{i=1}^{n_k}$, $\{ x^f_i\}_{i=1}^{n_f}$ and $\{ x^b_i\}_{i=1}^{n_b}$,  where $n_k$, $n_u$, $n_f$, and $n_b$ are the respective numbers of sensors. Here, one snapshot represents a simultaneous observation of all the sensors, and we assume that the data in one snapshot correspond to the same random event $\omega$ in the random space $\Omega$, while $\omega$ varies across different snapshots. 

Suppose we have a group of $N$ snapshots, denoted  by $\{ S(\omega^{(j)})\}^N_{j=1}$, defined as:
\begin{align}
\begin{split}
    \{S(\omega^{(j)})\}^N_{j=1} &= \{K(\omega^{(j)}), F(\omega^{(j)}), B(\omega^{(j)})\}^N_{j=1}, \\
    K(\omega^{(j)}) &= (k(x_i^k; \omega^{(j)}))_{i=1}^{n_k}, \\
    F(\omega^{(j)}) &= (f(x_i^f; \omega^{(j)}))_{i=1}^{n_f}, \\
    B(\omega^{(j)}) &= (b(x_i^b; \omega^{(j)}))_{i=1}^{n_b}. 
    \end{split}
\end{align}
We assume that we always have a sufficient number of sensors for the forcing term $f (x,  \omega)$ because accurate estimation of the forcing term is critical in physics-based training. 

\begin{figure}
  \centering
  \begin{tikzpicture}
    \draw[rounded corners, fill=anti-flashwhite] (0,3) rectangle (2,5);
    \node[text width=2cm,align=center] at (1.,4) {\scriptsize Real\\Snapshots\\$K(w^{(j)})$\\$F(w^{(j)})$\\$B(w^{(j)})$};
    \draw [-stealth](2,4) -- (2.5,4);
    \draw[fill=almond] (2.5,3) -- (2.5,5) -- (3.7,4.5) -- (3.7,3.5) -- cycle;
    \node[text width=4cm,align=center] at (3.1,4) {\scriptsize Encoder};
    \draw [-stealth](3.7,4) -- (4.2,4);

    \draw[rounded corners, fill=anti-flashwhite] (4.2,3) rectangle (6,5);
    \node[text width=4cm,align=center] at (5.1,4) {\scriptsize Latent\\variables\\$z^{(j)}$};

    \draw[rounded corners, fill=anti-flashwhite, line width=2, dashed] (4.2,0) rectangle (6,2);
    \node[text width=4cm,align=center] at (5.1,1) {\scriptsize Sampled\\variables\\${\zeta}^{(j)}$};
    \draw [-stealth,](5.1,2.5) -- (5.1,3);
    \draw [-stealth,](5.1,2.5) -- (5.1,2);
    \draw (5.1, 2.5) -- (7, 2.5);
    \draw (7, 2.5) -- (7, 0);
    \draw [-stealth, line width=2, dashed](6,1) -- (6.7,4);

    \draw[rounded corners, fill=anti-flashwhite, line width=2] (4.2,6) rectangle (6,8);
    \node[text width=4cm,align=center] at (5.1,7) {\scriptsize Coordinates\\$x_i$};
    \draw [-stealth,](6,7) -- (6.7,7);
    \draw [-stealth,](6,4) -- (6.7,4);
    \draw [-stealth, line width=2](6,7) -- (6.7,4);
    \draw [-stealth](6,4) -- (6.7,7);
    \draw[fill=almond, line width=2] (6.7,3.5) -- (6.7,4.5) -- (8.2,5) -- (8.2,3) -- cycle;
    \node[text width=4cm,align=center] at (7.45,4) {\scriptsize Decoder\\for u};
    \draw[fill=almond] (6.7,6.5) -- (6.7,7.5) -- (8.2,8) -- (8.2,6) -- cycle;
    \node[text width=4cm,align=center, line width=2] at (7.45,7) {\scriptsize Decoder\\for k};
    \draw [-stealth](8.2,7) -- (9,5.5);
    \draw [-stealth](8.2,4) -- (9,5.5);
    \draw [-stealth](8.2,7) -- (9,3.5);
    \draw [-stealth](8.2,4) -- (9,3.5);
    \draw [fill=anti-flashwhite] (9.5, 5.5) circle (0.5);
    \node[text width=4cm,align=center] at (9.5,5.5) {\scriptsize $\mathcal{N}_x$};
    \draw [fill=anti-flashwhite] (9.5, 3.5) circle (0.5);
    \node[text width=4cm,align=center] at (9.5,3.5) {\scriptsize $\mathcal{B}_x$};
    \draw[rounded corners, fill=anti-flashwhite] (10.5,6.5) rectangle (12,7.9);
    \node[text width=4cm,align=center] at (11.25,7.2) {\scriptsize Generated\\Snapshots\\$\tilde{K}(z^{(j)})$};
    
    \draw[rounded corners, fill=anti-flashwhite] (10.5,4.8) rectangle (12,6.2);
    \node[text width=4cm,align=center] at (11.25,5.5) {\scriptsize Generated\\Snapshots\\$\tilde{F}(z^{(j)})$};
    
    \draw[rounded corners, fill=anti-flashwhite] (10.5,2.8) rectangle (12,4.2);
    \node[text width=4cm,align=center] at (11.25,3.5) {\scriptsize Generated\\Snapshots\\$\tilde{B}(z^{(j)})$};
    
    \draw[rounded corners, fill=anti-flashwhite, line width=2, dashed] (10.5,0.5) rectangle (12,1.9);
    \node[text width=4cm,align=center] at (11.25,1.2) {\scriptsize Predicted\\response\\$\tilde{U}(z^{(j)})$};
    
    \draw [-stealth](8.2,7) -- (10.5,7.2);
    \draw [-stealth](10,3.5) -- (10.5, 3.5);
    \draw [-stealth](10,5.5) -- (10.5,5.5);
    \draw [-stealth, line width=2, dashed](8.2,4) -- (10.5,1.2);
    
    \draw (12,8) -- (12.2,8);
    \draw (12.2,8) -- (12.2, 2.7);
    \draw (12, 2.7) -- (12.2, 2.7);
    \draw (12.2, 5.35) -- (12.4, 5.35);
    
    \draw[rounded corners, fill=anti-flashwhite] (12.5,4.35) rectangle (14.5,6.35);
    \node[text width=4cm,align=center] at (13.5, 5.35) {\scriptsize Reconstructed\\Snapshots\\$\tilde{K}(z^{(j)})$\\$\tilde{F}(z^{(j)})$\\$\tilde{B}(z^{(j)})$};
    
    \draw[rounded corners, fill=anti-flashwhite] (6,-1) rectangle (8,0);
    \node[text width=4cm,align=center] at (7, -0.5) {\scriptsize MMD-based\\Loss};
    
    \draw (1,-0.5) -- (6,-0.5);
    \draw (8,-0.5) -- (13.5,-0.5);
    \draw [-stealth](1,-0.5) -- (1,3);
    \draw [-stealth](13.5,-0.5) -- (13.5, 4.35);
    
  \end{tikzpicture}
  \centering
  \caption{\footnotesize The architecture of the PI-VAE framework for solving forward problems. Thin lines show the parts used only during training; solid thick lines show the parts used both in training and test stages; dashed thick lines are the parts used only during test.}
  \label{fig.scheme_forward}
  
  \end{figure}
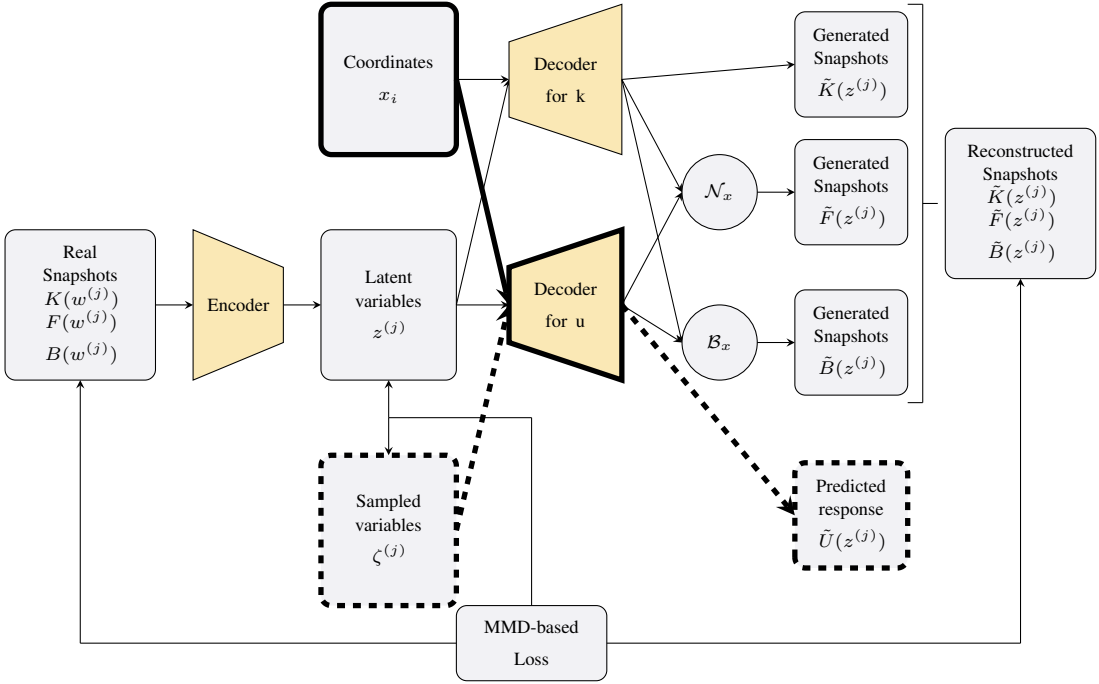

The solution of the SDE is calculated by constructing the PI-VAE framework via the following three steps:
\begin{itemize}
  \item Firstly, we use fully connected feed-forward neural networks as our encoder and decoders. An encoder $\mathcal{E}_{\phi}$ maps input data to the latent variables $z$, i.e. $z^{(j)} = \mathcal{E}_{\phi}(S(\omega^{(j)}))$. Two independent decoders $\tilde{k}_{\theta_k}(x,z)$ and $\tilde{u}_{\theta_u}(x,z)$ approximate corresponding stochastic process $k(x,\omega)$ and $u(x,\omega)$ respectively by constructing a mapping from coordinates $x$ and latent variable $z$ to input data. We expect to generate approximate  QoIs using  the well-trained decoders after the training process. 
  
  \item Second, inspired by physics-informed neural network for deterministic differential equations \cite{raissi2017physics,raissi2017physics1}, we encode the governing differential equation into the framework by applying operator $\mathcal{N}_x$ and $\mathcal{B}_x$ on the decoder outputs,  $\tilde{k}_{\theta_k}$ and $\tilde{u}_{\theta_u}$, to obtain approximations of the $f(x, \omega)$ and $b(x, \omega)$ in the governing SDE, i.e.:
  \begin{align*}
    \tilde{f}_{\theta_u,\theta_k}(x, z) &= \mathcal{N}_x [ \tilde{u}_{\theta_u}(x,z), \tilde{k}_{\theta_k}(x,z)],\\
    \tilde{b}_{\theta_u}(x,z) &= \mathcal{B}_x [ \tilde{u}_{\theta_u}(x,z)].
  \end{align*}
  Differentiations in $\mathcal{N}_x$ and  $\mathcal{B}_x$ are carried out by automatic differentiation \cite{paszke2017automatic}. These physics-informed estimates $\tilde{f}_{\theta_u,\theta_k}(x, z)$ and $\tilde{b}_{\theta_u}(x, z)$ together with approximated response  $\tilde{u}_{\theta_u}(x,z)$ and system parameters $\tilde{k}_{\theta_k}(x,z)$ constitutes  the reconstructed snapshots $\{\tilde{S}(z^{(j)})\}$, i.e.
    \begin{align*}
  \{\tilde{S}(z^{(j)})\}_{j=1}^N &= \{\tilde{K}(z^{(j)}), \tilde{F}(z^{(j)}), \tilde{B}(z^{(j)})\}_{j=1}^N, \\
  \tilde{K}(z^{(j)}) &= (\tilde{k}_{\theta_k}(x_i^k ,z^{(j)}))_{i=1}^{n_k}, \\
   \tilde{F}(z^{(j)}) &= (\tilde{f}_{\theta_u, \theta_k}(x_i^f ,z^{(j)}))_{i=1}^{n_f}, \\
  \tilde{B}(z^{(j)}) &= (\tilde{b}_{\theta_u}(x_i^b ,z^{(j)}))_{i=1}^{n_b}\},
  \end{align*}

  \item In the third step, we formulate the loss function which consists of the reconstruction term and a regularization term. The reconstruction cost is calculated using the MMD between collected sensor measurements $\{S(\omega^{(j)})\}_{j=1}^N$ and reconstructed samples $\{\tilde{S}(z^{(j)})\}_{j=1}^N$. The regularization term promotes the distribution of the latent encodings to be close to standard normal distribution. This term measures the MMD between latent variables $z$ (obtained by encoding the measured snapshots) and  random variables $\zeta$ distributed according to the prior distribution, here chosen to be standard normal. Thus, the loss function is given by 

  \[
    L(\phi, \theta_u, \theta_k)  = \sum_{i=1}^M \text{MMD}_{k_i,U}(\tilde{S}(z), S(\omega)) + \sum_{i=1}^M \text{MMD}_{k_i,U}(z,\zeta). 
  \]
    
 After training using this loss function, the decoders $\tilde{u}_{\theta_u}(x;\zeta)$ and $\tilde{k}_{\theta_k}(x;\zeta)$ is used to approximate the stochastic process $u(x,\omega)$ and $k(x,\omega)$ at various locations $x$. A detailed description of our training algorithm is provided in Algorithm \ref{alg.forward} and Figure \ref{fig.scheme_forward}. 
  
\end{itemize}

\begin{algorithm}
    \SetAlgoLined
    Initialization: Set the training steps $n_t$, batch size $n$, Adam hyper-parameters $\alpha, \beta_1, \beta_2$, initial values for encoder and decoder parameters   $\phi, \theta_u, \theta_k$, and kernels of MMD estimator $\{k_i\}_{i=1}^M$. \\
    \For {$i = 1,2,...,n_t$}
    {
    Sample $N$ snapshots $\{S(\omega^{(j)})\}_{j=1}^N$. \\
    Sample $N$ random vectors from the prior $\{\zeta_j\}_{j=1}^N$ . \\
    \For{$j = 1,2,...,N$}
      {
        $z_j = \mathcal{E}_{\phi}(S(\omega^{(j)}))$ \\
        \{$\tilde{K}(z^{(j)}), \tilde{F}(z^{(j)}), \tilde{B}(z^{(j)})\} = \{k_{\theta_k}(x, z^{(j)}), \tilde{f}_{\theta_u, \theta_k}(x, z^{(j)}), \tilde{b}_{\theta_u} (x, z^{(j)})$\} \\
        $\tilde{S}(z^{(j)}) = [\tilde{K}(z^{(j)}),\tilde{F}(z^{(j)}),\tilde{B}(z^{(j)})]$
      }
    $L = \sum_i^M   \text{MMD}_{k_i,U}( \{\tilde{S}(z^{(j)})\}_{j=1}^N,  \{{S}(\omega^{(j)})\}_{j=1}^N) + \text{MMD}_{k_i,U}( \{z^{(j)}\}_{j=1}^N, \{\zeta^{(j)}\}_{j=1}^N) $ \\
    $\phi, \theta_u, \theta_k \leftarrow \text{Adam}(\nabla L, \phi, \theta_u, \theta_k, \alpha, \beta_1, \beta_2 )$
    }
   \caption{PI-VAE for solving SDEs (forward problems)}
   \label{alg.forward}
  \end{algorithm}

\subsection{Solving inverse problems}
In inverse problems, the goal is to calculate the unknown model parameters using measurement data from the response and boundary conditions. Following the formulation~\ref{eq.sde}, we seek to estimate the random parameter $k(x,\omega)$ using measurements from scattered sensors  on these processes, $u(x,\omega)$, $f(x,\omega)$ and $b(x,\omega)$, at locations , $\{ x^u_i\}_{i=1}^{n_u}$, $\{ x^f_i\}_{i=1}^{n_f}$ and $\{ x^b_i\}_{i=1}^{n_b}$,  where $n_u$, $n_f$, and $n_b$ are the respective numbers of sensors. Since we have a physics-based loss which involves calculation of the residual, we also add a few measurements of $k(x,\omega)$ at $\{ x^k_i\}_{i=1}^{n_k}$ with $n_k << n_u$.  Suppose we have a group of $N$ snapshots, denoted  by $\{ S(\omega^{(j)})\}^N_{j=1}$, defined as:
\begin{align}
\begin{split}
    \{S(\omega^{(j)})\}^N_{j=1} &= \{K(\omega^{(j)}), U(\omega^{(j)}), F(\omega^{(j)}), B(\omega^{(j)})\}^N_{j=1}, \\
    K(\omega^{(j)}) &= (k(x_i^k; \omega^{(j)}))_{i=1}^{n_k}, \\
    U(\omega^{(j)}) &= (u(x_i^u; \omega^{(j)}))_{i=1}^{n_u}, \\
    F(\omega^{(j)}) &= (f(x_i^f; \omega^{(j)}))_{i=1}^{n_f}, \\
    B(\omega^{(j)}) &= (b(x_i^b; \omega^{(j)}))_{i=1}^{n_b}. 
    \end{split}
\end{align}

\begin{figure}
  \centering
  \begin{tikzpicture}
    \draw[rounded corners, fill=anti-flashwhite] (0,3) rectangle (2,5);
    \node[text width=2cm,align=center] at (1.,4) {\scriptsize Real\\Snapshots\\$K(w^{(j)})^*$\\$F(w^{(j)})$\\$B(w^{(j)})$};
    \draw [-stealth](2,4) -- (2.5,4);
    \draw[fill=almond] (2.5,3) -- (2.5,5) -- (3.7,4.5) -- (3.7,3.5) -- cycle;
    \node[text width=4cm,align=center] at (3.1,4) {\scriptsize Encoder};
    \draw [-stealth](3.7,4) -- (4.2,4);
    \draw[rounded corners, fill=anti-flashwhite] (4.2,3) rectangle (6,5);
    \node[text width=4cm,align=center] at (5.1,4) {\scriptsize Latent\\variables\\$z^{(j)}$};

    \draw[rounded corners, fill=anti-flashwhite, line width=2, dashed] (4.2,0) rectangle (6,2);
    \node[text width=4cm,align=center] at (5.1,1) {\scriptsize Sampled\\variables\\${\zeta}^{(j)}$};
    \draw [-stealth,](5.1,2.5) -- (5.1,3);
    \draw [-stealth,](5.1,2.5) -- (5.1,2);
    \draw (5.1, 2.5) -- (7, 2.5);
    \draw (7, 2.5) -- (7, -1);
    \draw [-stealth, line width=2, dashed](6,1) -- (6.7,7);

    \draw[rounded corners, fill=anti-flashwhite, line width=2] (4.2,6) rectangle (6,8);
    \node[text width=4cm,align=center] at (5.1,7) {\scriptsize Coordinates\\$x_i$};
    \draw [-stealth, line width=2](6,7) -- (6.7,7);
    \draw [-stealth](6,4) -- (6.7,4);
    \draw [-stealth](6,7) -- (6.7,4);
    \draw [-stealth,](6,4) -- (6.7,7);
    \draw[fill=almond] (6.7,3.5) -- (6.7,4.5) -- (8.2,5) -- (8.2,3) -- cycle;
    \node[text width=4cm,align=center] at (7.45,4) {\scriptsize Decoder\\for u};
    \draw[fill=almond, line width=2] (6.7,6.5) -- (6.7,7.5) -- (8.2,8) -- (8.2,6) -- cycle;
    \node[text width=4cm,align=center, line width=2] at (7.45,7) {\scriptsize Decoder\\for k};

    \draw [-stealth](8.2,7) -- (9,3.5);
    \draw [-stealth](8.2,4) -- (9,3.5);

    \draw [fill=anti-flashwhite] (9.5, 3.5) circle (0.5);
    \node[text width=4cm,align=center] at (9.5,3.5) {\scriptsize $\mathcal{N}_x$};
    \draw[rounded corners, fill=anti-flashwhite, line width=2] (10.5,6.5) rectangle (12,7.9);
    \node[text width=4cm,align=center] at (11.25,7.2) {\scriptsize Generated\\Snapshots\\$\tilde{K}(z^{(j)})$};
    
    \draw[rounded corners, fill=anti-flashwhite] (10.5,4.8) rectangle (12,6.2);
    \node[text width=4cm,align=center] at (11.25,5.5) {\scriptsize Generated\\Snapshots\\$\tilde{U}(z^{(j)})$};
    
    \draw[rounded corners, fill=anti-flashwhite] (10.5,2.8) rectangle (12,4.2);
    \node[text width=4cm,align=center] at (11.25,3.5) {\scriptsize Generated\\Snapshots\\$\tilde{F}(z^{(j)})$};
    
    \draw [-stealth, line width=2](8.2,7) -- (10.5,7.2);
    \draw [-stealth](10,3.5) -- (10.5,3.5);
    \draw [-stealth](8.2,4) -- (10.5,5.5);
    
    \draw (12,8) -- (12.2,8);
    \draw (12.2,8) -- (12.2, 2.7);
    \draw (12, 2.7) -- (12.2, 2.7);
    \draw (12.2, 5.35) -- (12.4, 5.35);
    
    \draw[rounded corners, fill=anti-flashwhite] (12.5,4.35) rectangle (14.5,6.35);
    \node[text width=4cm,align=center] at (13.5, 5.35) {\scriptsize Reconstructed\\Snapshots\\$\tilde{K}(z^{(j)})$\\$\tilde{F}(z^{(j)})$\\$\tilde{B}(z^{(j)})$};
    
    \draw[rounded corners, fill=anti-flashwhite] (6,-1.5) rectangle (8,-0.5);
    \node[text width=4cm,align=center] at (7, -1) {\scriptsize MMD-based\\Loss};
    
    \draw (1,-1) -- (6,-1);
    \draw (8,-1) -- (13.5,-1);
    \draw [-stealth](1,-1) -- (1,3);
    \draw [-stealth](13.5,-1) -- (13.5, 4.35);
    
  \end{tikzpicture}
  \centering
  \caption{\footnotesize The architecture of the PI-VAE framework for solving mixed and inverse problems. Thin lines show the parts used only during training; solid thick lines show the parts used both in training and test stages; dashed thick lines are the parts used only during test. Note that only few samples from $k$ is used in inverse and mixed problems.}
  \label{fig.scheme_inverse}

\end{figure}
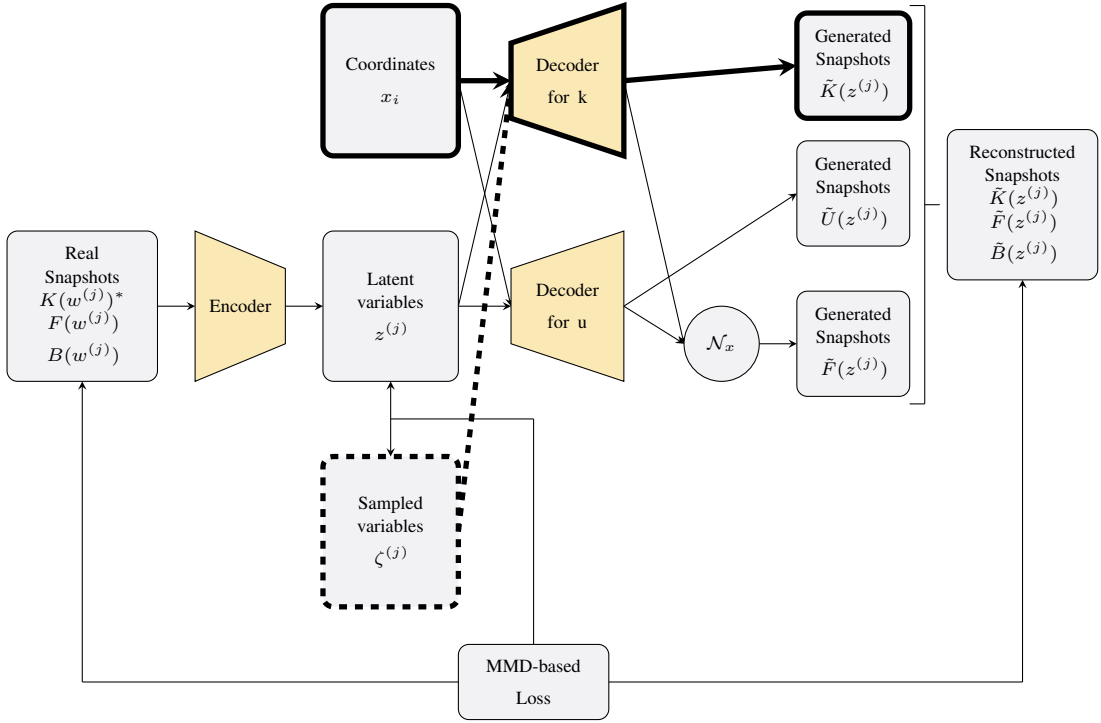

We assume that we always have sufficient number of sensors for the forcing term $f (x,  \omega)$ because accurate estimation of the forcing term is critical in physics-based training. We take similar steps to the three steps discussed in the previous section and solve the inverse and mixed problems according to Algorithm \ref{alg.inverse} and Figure \ref{fig.scheme_inverse}.

\begin{algorithm}
    \SetAlgoLined
    Initialization: Set the training steps $n_t$, batch size $n$, Adam hyper-parameters $\alpha, \beta_1, \beta_2$, initial values for encoder and decoder parameters   $\phi, \theta_u, \theta_k$, and kernels of MMD estimator $\{k_i\}_{i=1}^M$. \\
    \For {$i = 1,2,...,n_t$}
    {
    Sample $N$ snapshots $\{S(\omega^{(j)})\}_{j=1}^N$. \\
    Sample $N$ random vectors from the prior $\{\zeta_j\}_{j=1}^N$ . \\
    \For{$j = 1,2,...,N$}
      {
        $z_j = \mathcal{E}_{\phi}(S(\omega^{(j)}))$ \\
        \{$\tilde{K}(z^{(j)}), \tilde{F}(z^{(j)}), \tilde{U}(z^{(j)})\} = \{k_{\theta_k}(x, z^{(j)}), \tilde{f}_{\theta_u, \theta_k}(x, z^{(j)}), u_{\theta_u} (x, z^{(j)})$\} \\
        $\tilde{S}(z^{(j)}) = [\tilde{K}(z^{(j)}),\tilde{F}(z^{(j)}),\tilde{U}(z^{(j)})]$
      }
    $L = \sum_i^M  \{\tilde{S}(z^{(j)})\}_{j=1}^N,  \{{S}(\omega^{(j)})\}_{j=1}^N) + \text{MMD}_{k_i,U}( \{z^{(j)}\}_{j=1}^N, \{\zeta^{(j)}\}_{j=1}^N) $ \\
    $\phi, \theta_u, \theta_k \leftarrow \text{Adam}(\nabla L, \phi, \theta_u, \theta_k, \alpha, \beta_1, \beta_2 )$
    }
   \caption{PI-VAE for solving mixed and inverse SDEs}
   \label{alg.inverse}
\end{algorithm}

\section{Numerical results} \label{sec.results}

In this section, we numerically evaluate how accurately the proposed VAE framework can approximate a stochastic process, and how well the proposed  PI-VAE approach can solve SDEs. We compare our results with Wasserstein GAN with gradient penalty (WGAN-GP) and Physics-informed WGAN (PI-WGANs) \cite{yang2018physics}. 

All test cases commonly share the following default settings unless mentioned otherwise. We use the hyperbolic tangent function (tanh) as our activation function to ensure the smoothness in high-order derivatives of the developed neural networks. Every feed-forward network (i.e. encoders and decoders of VAE; generator and discriminator of WGAN) has four hidden layers of width 128. The latent space is chosen to be an independent multivariate standard Gaussian multivariate. Adam is the default optimizer with the following default hyper-parameters: $\beta_1$ = 0.5 and $\beta_2$ = 0.9. For WGAN-GP, we set the hyper-parameters in the loss function  $\lambda$=0.1, $n_d$=5 as default values similarly to \cite{yang2018physics}. For the hyper-parameter of Sinkhorn Iteration, we set the default value $\epsilon$=0.1. For the basic setting of MMD, we will use five radial basis function (RBF) kernels of different bandwidths for calculation. In all numerical experiments of approximating stochastic processes, the learning rate is 0.001 as the default value. For  SDE cases, we consider special settings for the learning rate, which will be explained in the following sections. We collect all the training data by numerically simulating sensor measurements of the stochastic processes based on  Finite Difference Method and Monte Carlo Simulation.

\subsection{Approximating stochastic processes}

\subsubsection{Processes with different kernels}

\begin{figure}
    \centering
    \includegraphics[width=15.5cm]{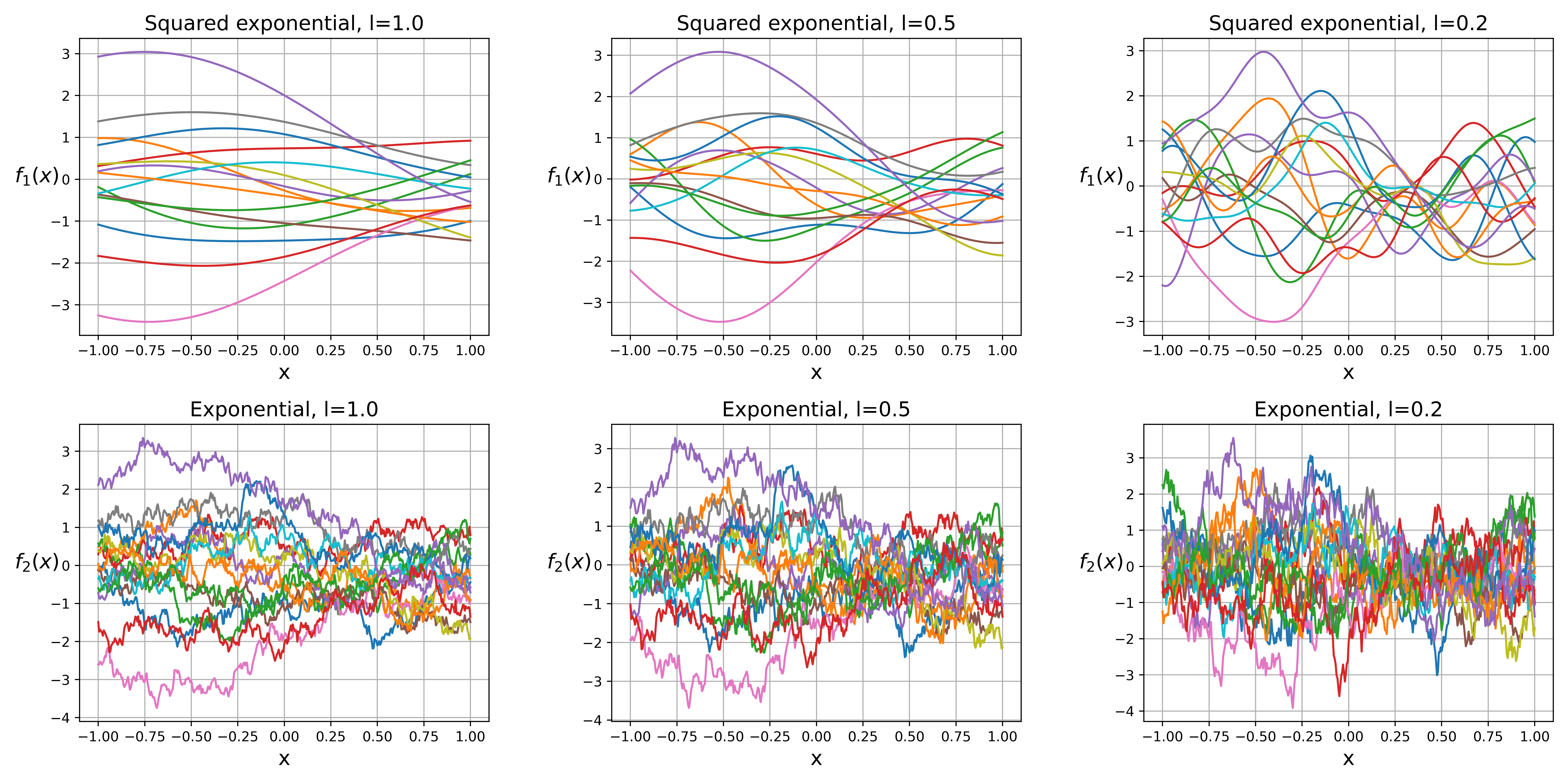}
    \centering
    \caption{\footnotesize Sample paths of Gaussian processes of different kernel function and length scale. Sample paths of different correlation lengths: l = 1 (left), 0.5 (middle), and 0.2 (right) and different kernels: squared exponential kernel (top) and exponential kernel (bottom) are shown.}
    \label{fig.GP_samples}
\end{figure}

We consider Gaussian Processes with zero mean and two different kernels: a  squared exponential kernel and an exponential kernel, as follows
\begin{align*}
  f_1(x) &\sim \text{GP} (0, \exp(-\frac{1}{2 l^2}(x-x')^2)),  &x,x' \in  [-1,1],\\
   f_2(x) &\sim \text{GP} (0, \exp(-\frac{1}{2 l^2}(x-x')^2)),  &x,x' \in [-1,1],
\end{align*}
where $l$ is the correlation length scale, which will be set to 0.2, 0.5, or 1. The number of sensors is 6 or 11. Sensors are uniformly placed over the domain $[-1 , 1]$. In the following case, to show that our method can find the pattern from a limited number of scattered measurements, we will only use 2,000 snapshots for our training data as the default setting. Also, to ensure that the effect of batch size is considered, we will use a batch size of 1000 for our training, which is different from \cite{yang2018physics}. For each case, we run the code with different random seeds. Few sample paths of the stochastic processes are shown in Figure \ref{fig.GP_samples}. As we can see in the figure, stochastic processes of the exponential kernel results in more locally correlated processes, which will be more challenging to be approximated. 

\begin{figure}
\centering
\includegraphics[width=15cm]{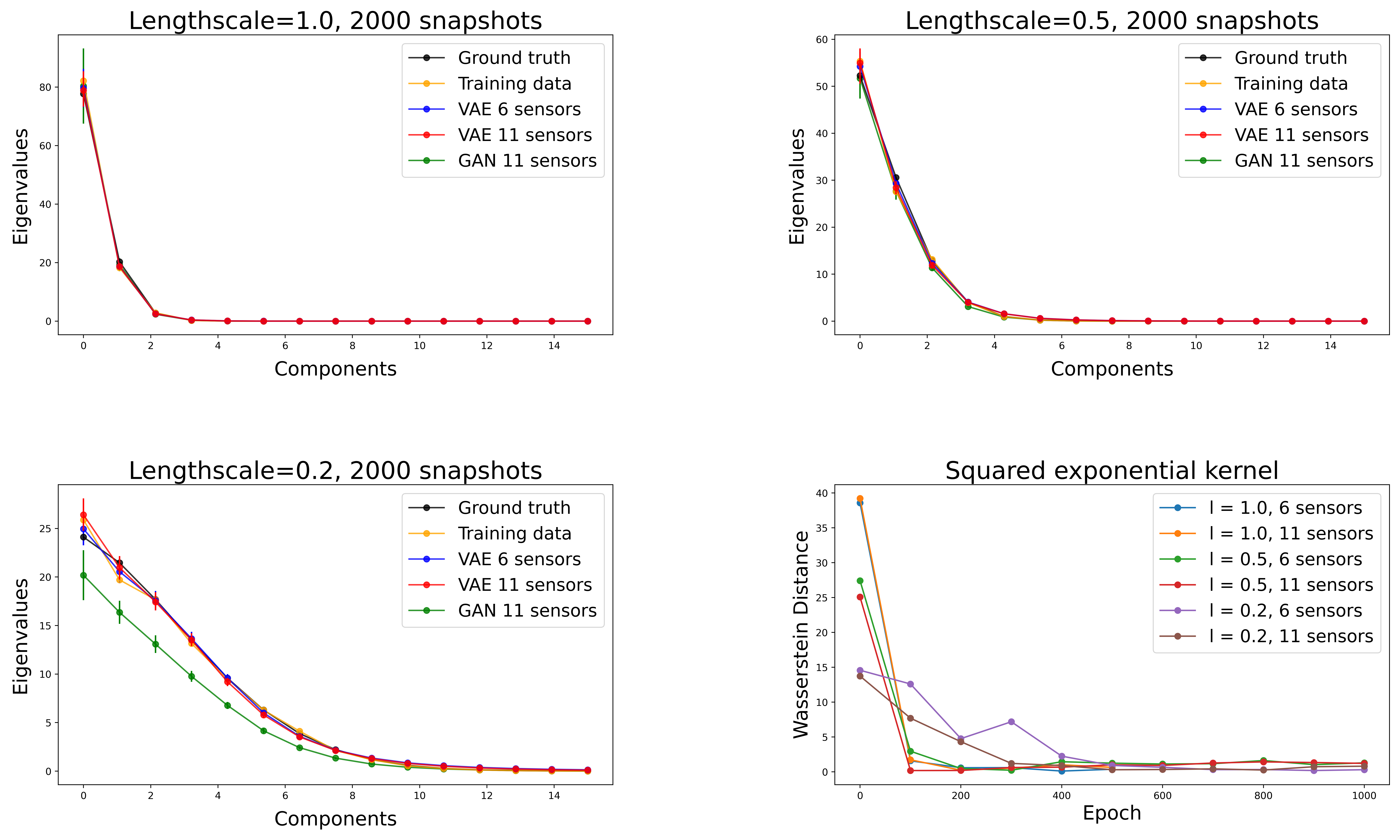}
\centering
\caption{\footnotesize Results of using our method to approximate Gaussian Process of squared exponential kernel.The first three plots shows the spectral of the covariance matrix of the reconstructed samples and ground true samples. The last plot shows the similarity of reconstructed samples and ground true samples in terms of $\bm{W_1}$ distance.}
\label{fig.square_exp_results}
\end{figure}

\begin{figure}
\centering
\includegraphics[width=16cm]{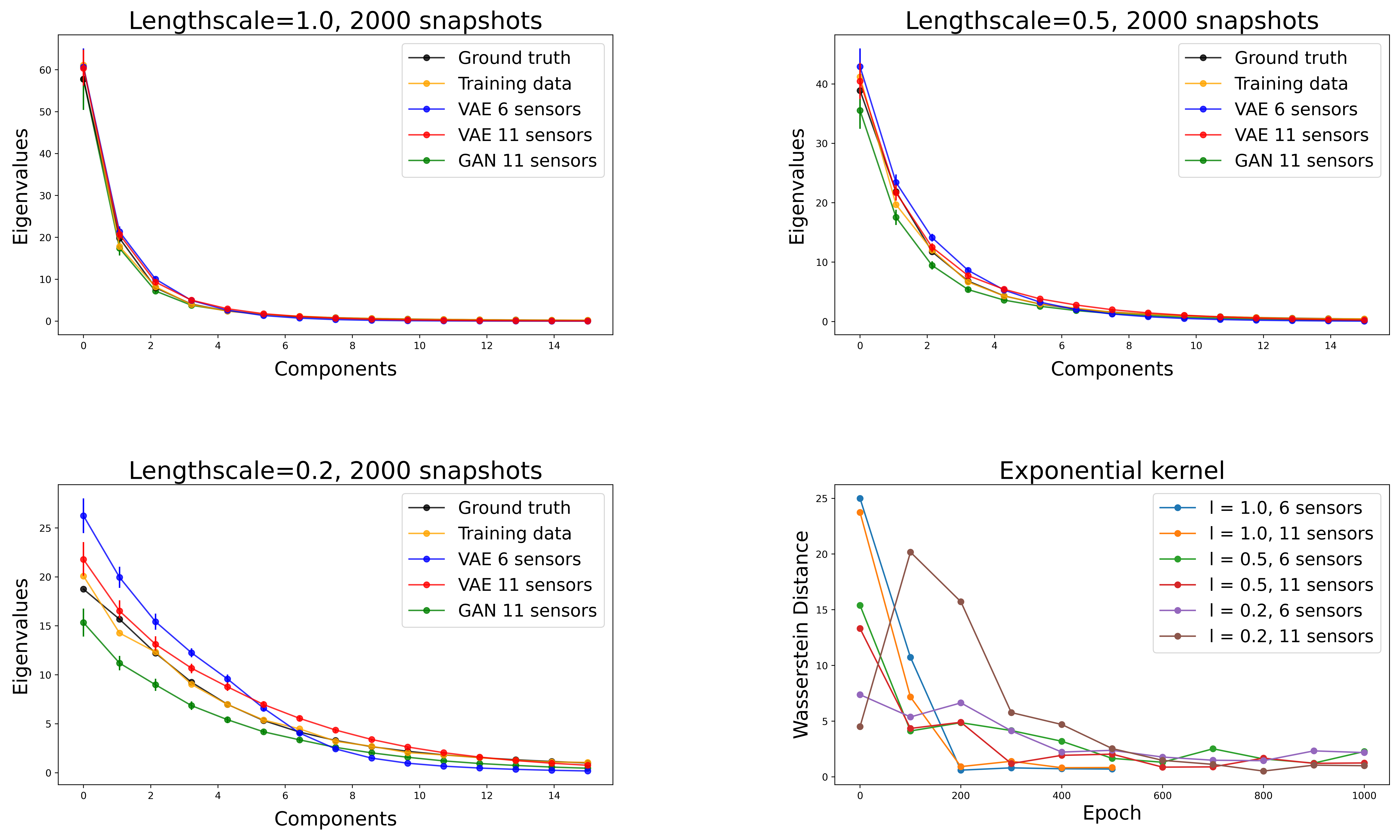}
\caption{\footnotesize Results of using our method to approximate Gaussian Process of exponential kernel.The first three plots shows the spectral of the covariance matrix of the reconstructed samples and ground true samples. The last plot shows the similarity of reconstructed samples and ground true samples in terms of $\bm{W_1}$ distance. }

\label{fig.exp_results}
\end{figure}

For length scales of 1.0, 0.5, and 0.2, we train our VAE model and WGAN-GP for 500, 1000, and 1500 training epochs, respectively.  After the training is finished, we will use 101 uniformly distributed coordinates over the domain for validation of the trained neural network. We pick 100 decoders at the last 100 training epochs and use them to reconstruct sample paths on the validation coordinates as reconstructed samples. The ground truth samples are collected by re-sampling of the stochastic processes. To validate the accuracy of the algorithms, in  Figures \ref{fig.square_exp_results} and  \ref{fig.exp_results}, respectively for squared exponential and exponential kernels, we compare the reconstructed and ground truth sample paths in terms of (1) the spectra of the sample paths, i.e., the eigenvalues of the covariance matrices, and (2) the Wasserstein distance $\bm{W_1}$.

In these figures, the $\bm{W_1}$ distance between reconstructed samples and ground true samples are shown at different training epochs, and one can see the convergence after 500 training epochs. It can be seen that the Wasserstein distance for all the cases  converge to approximately 1, except for exponential kernels with length scales of 0.5 and 0.2, where we cannot achieve the same desired accuracy when we  use only 6 sensors. 

To compare the eigenvalues, as the ground truth, we calculate the eigenvalues of the analytical form of the covariance matrix of the stochastic processes. The eigenvalues of the reconstructed sample paths are calculated using 1,000 training sample paths, and as can be seen from the the spectra of the covariance matrices of these process, our model can effectively reconstruct samples with similar spatial distributions. It should be noted that the approximation errors is larger for the more dominant eigenvectors. Also, the approximation errors are smaller in the squared exponential kernel case  since these processes are less locally correlated, compared to processes with exponential kernels, and are easier to approximate. Similar justification can be used for the larger error observed in smaller  length scales.  Improvements in approximation errors can be seen when more sensors are used. In the cases of length scale = 1.0 and 0.5, WGAN and our method can accurately reproduce the samples of covariance of similar spectral structures. However, for  the smaller length scale of 0.2, our VAE model has slightly better performance compared to WGAN-GP.

\subsubsection{Different choices of loss functions}

As mentioned  in Section \ref{sec.method_GP}, the VAE model can be built using  different  distance measures between probability distribution. In this section, we evaluate the performance of our proposed model  when different distance measures are used, we now use different distance measures as the loss function in the training of (unconstrained) Gaussian processes and  Gaussian processes constrained on the boundaries. 

As mentioned before, the loss function of the VAE consists of dissimilarity measures in both latent space and input data space. In this work, we  always use MMD as the dissimilarity measure in the latent space. In the high-dimensional data space, we consider three choices for the similarity measure: Mean Square Error (MSE), Sinkhorn Iteration and MMD, and assess the three corresponding models. Although cross entropy loss is also a popular loss function, it is  recommended to adopt MSE as the loss function when the data of original space is not binary \cite{kingma2013auto}.

\begin{figure}
    \includegraphics[width=16cm]{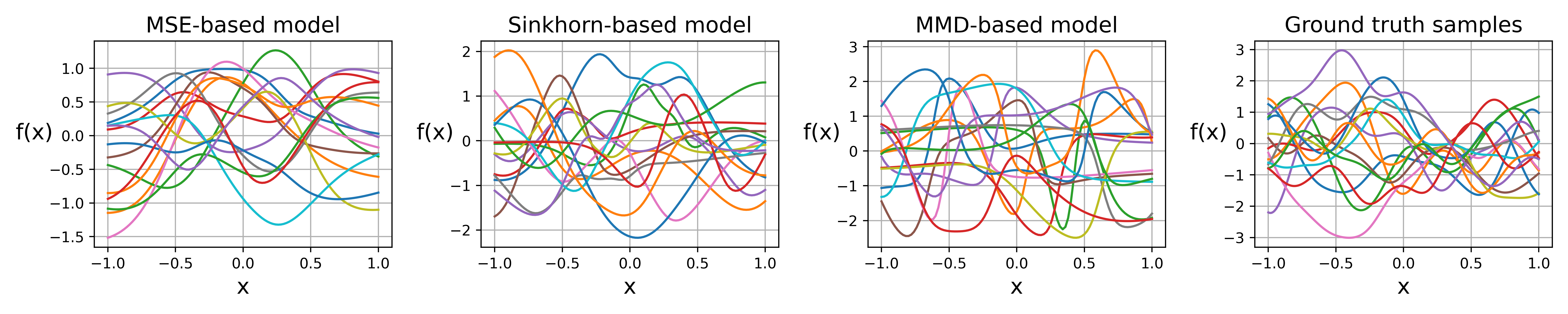}
    \caption{\footnotesize Reconstructed samples of models trained by different loss function which are MSE, Sinkhorn and MMD in input space. The ground truth samples are also shown for comparison. }
    \label{fig.diff_loss_samples}
\end{figure}

The reconstructed samples of models of different loss functions are shown in Figure \ref{fig.diff_loss_samples}. All these models can reproduce smooth sample paths, which means that functions represented by our trained model have good continuity. Compared with ground truth samples, samples of MSE loss are much more spatially concentrated than the other two models. Although samples from the model using  Sinkhorn loss are similar to the training sample paths, the length scale of the sample paths is not as small as the training samples. Among these three models, the model with the MMD loss can reproduce  samples that are most similar to the training data.

\begin{figure}
    \centering
    \includegraphics[width=14cm]{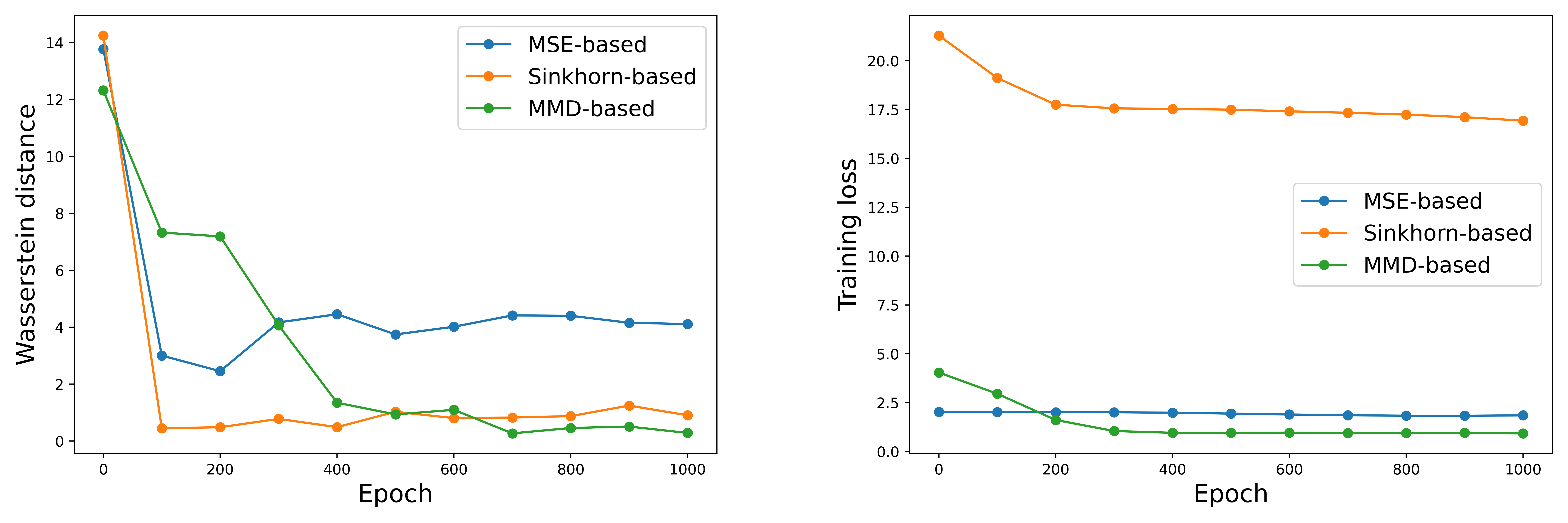}
    \caption{\footnotesize Wasserstein distance (left) and training loss (right) versus the training epoch of the three models using different training loss.}
    \label{fig.diff_loss_wd&loss}
\end{figure}

The following numerical results are in accordance with the results shown in Figure \ref{fig.diff_loss_wd&loss}, which shows the training loss  converges  and that there is no over-fitting. But we observe that the Wasserstein distance from the MSE-based VAE model is significantly larger than that from the other two models, which indicates that the generated samples from the MSE-based VAE model does not match the distribution of input data statistically. Besides, the Wasserstein distance for the Sinkhorn-based model is slightly larger than  the Wasserstein distance of the MMD-based model. 

\begin{figure}
    \centering
    \includegraphics[width=8cm]{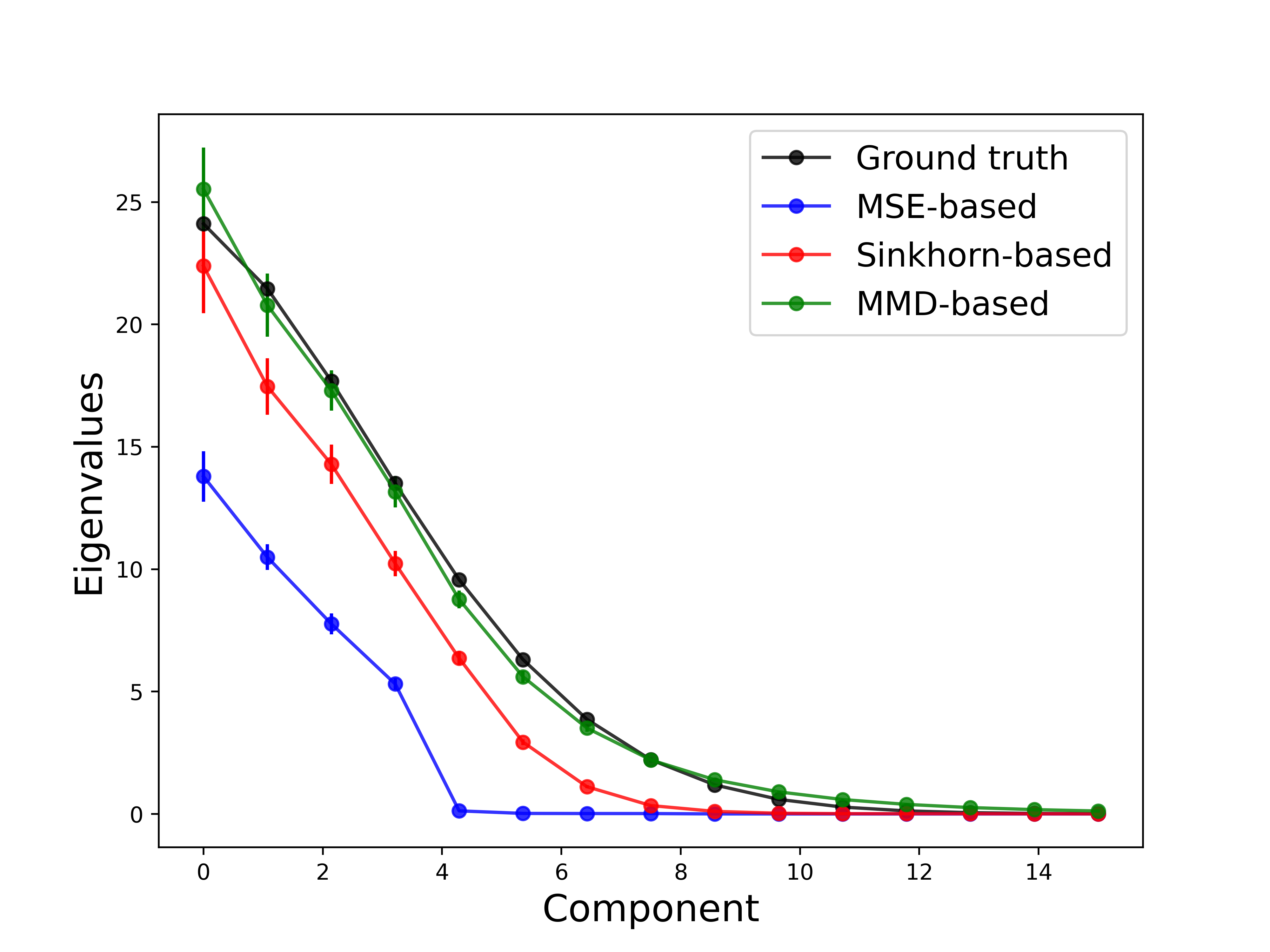}
    \caption{\footnotesize the spectral of the covariance matrix of the reconstructed samples of different models and ground truth samples}
    \label{fig.diff_loss_cov}
\end{figure}

The spectra comparison in Figure \ref{fig.diff_loss_cov}  shows that the Sinkhorn-based model and MSE-based models fail to capture covariance information as accurately as the MMD-based model. The MSE-based model can not perform as well as the other two models. This can be due to the fact that the MSE is only verified to be effective when input data follows a Gaussian distribution with an approximately diagonal covariance \cite{kingma2013auto}. For the case of highly correlated Gaussian data, it seems that MSE cannot give us a very promising results. However, it is surprising to observe that Sinkhorn model performs worse than MMD model because theoretically Sinkhorn divergence is closer to Wasserstein distance than MMD \cite{patrini2020sinkhorn}. We can associate this to the suboptimal selection of the hyper-parameter of Sinkhorn algorithm. 

One of the drawbacks  of Sinkhorn Iteration is  the difficulty in choosing an appropriate hyper-parameter, specifically the convergence threshold \cite{xie2020fast}. In our numerical experiments, we  observed that the whole training will easily collapse when a small threshold is used, while a  large threshold leads to inaccurate estimation of the distribution distance\cite{xie2020fast}. So, we note that the Sinkhorn-based model can be potentially improved with further fine-tuning of its hyper-parameter, but also underscore that the MMD-based model is the most advantageous since it is the most accurate model without any need for hyper-parameter fine-tuning.

Let us now approximate a stochastic process with a fixed boundary condition. Specifically, we consider $f(x, \omega) = (x^2 - 1)g(x, \omega)$, where $g(x, \omega)$ is a Gaussian process. The mean and standard deviation of the generated samples are compared for accuracy evaluation. We use 100 decoders and evaluate their average performance after 500 training epochs. It is observed that Sinkhorn-based model can approximate better on the boundary than MMD model, which indicates that Sinkhorn model can better detect the pattern of concentrated data distribution. 

\begin{figure}
    \flushleft
    \includegraphics[width=16cm]{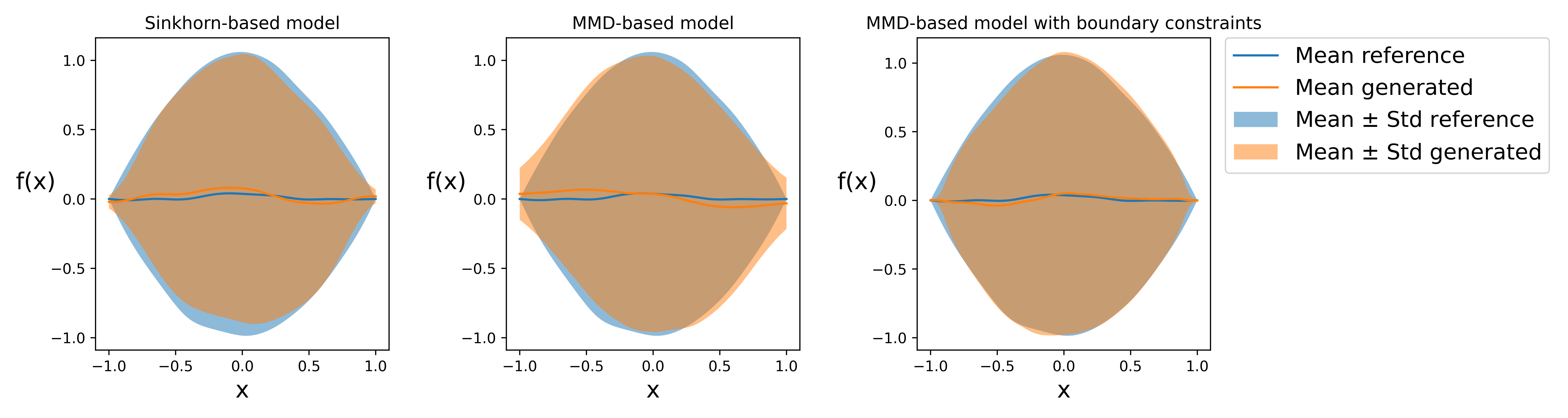}
    \caption{\footnotesize Mean and variance estimate of Sinkhorn-based model (left) and MMD-based model (mid) and the MMD-based model with MSE penalty on the boundary (right). Blue line represent the reference mean and blue shade area represents the reference standard deviation. In contrast, orange line and orange shade area represent the mean and std of the reconstructed samples, respectively.}
    \label{fig.low_manifold_std_plot}
\end{figure}

However, in some cases (e.g. solving SDEs), we will have an additional penalty term in the loss function to constrain the boundary condition. To increase the accuracy of our approximation on the boundary, we can add a penalty term on the boundary using MSE loss for MMD-based model. In our example, no trade-off is needed between different loss terms. As we can see in Figure \ref{fig.low_manifold_std_plot}, the approximation is much better than the vanilla MMD model.

\subsubsection{Issue of mode collapse}
The MMD measure calculates the  distance between probability distributions using finite samples of reconstruction  and input data. The number of samples will affect the accuracy of the estimation. Theoretically, the number of samples needed for MMD estimation is approximately proportional to the dimensionality of the data space \cite{reddi2015high}. This indicates that MMD estimation will not suffer from the "curse of dimensionality". In the numerical examples presented in this paper, the dimensions of the data are not large, allowing us to use regular mini-batch sizes to measure MMD. However, it is still necessary to assess the effect of batch size on  the performance of the proposed method. 

\begin{figure}
    \centering
    \includegraphics[width=14cm]{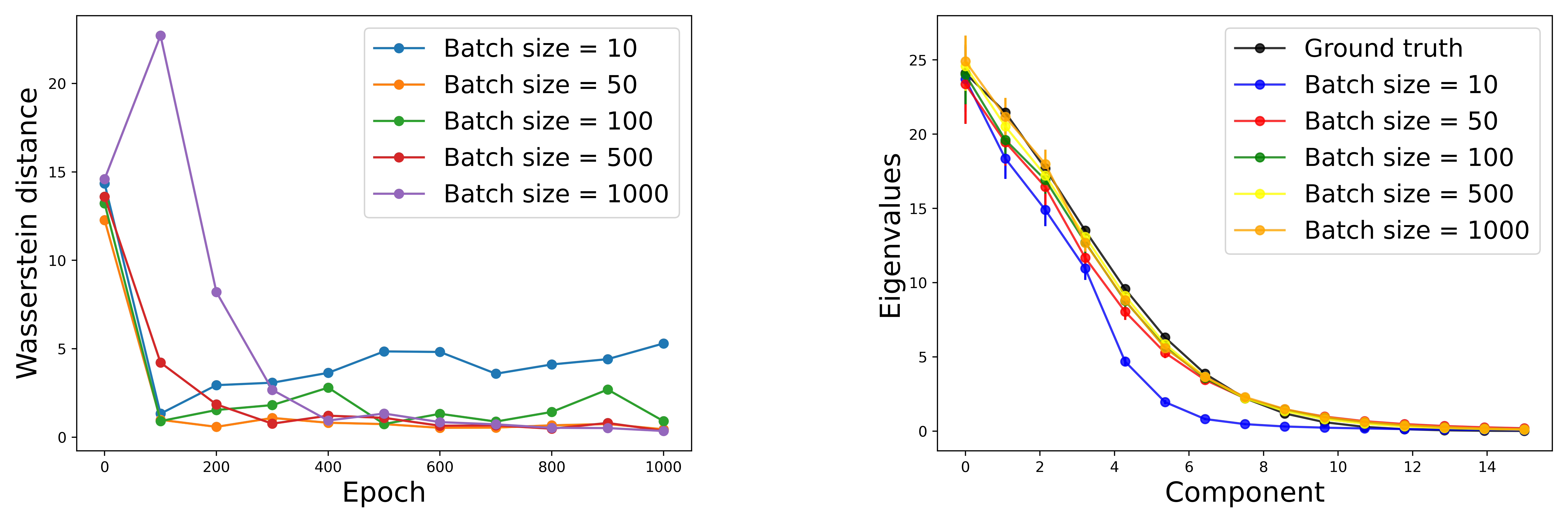}
    \caption{\footnotesize Wasserstein distance (left) of reconstructed samples and ground true samples and spectral of the covariance matrix (rigth) of the reconstructed samples of different batch size are shown.}
    \label{fig.batchsize_effect_wd&cov}
\end{figure}

To this end, we show the calculated Wasserstein distance and the spectra of the covariance matrix   in Figure \ref{fig.batchsize_effect_wd&cov} for different batch sizes. It can be seen the performance of our proposed model will decrease as the batch size in each training epoch decreases. This indicates that the reduction in the number of samples deteriorates the accuracy in MMD estimation. However, this performance decline in the covariance spectra is not significant except for  the batch size of 10. Similar observation can be made for the Wasserstein distance results, showing that our model is not sensitive to the common choices of batch size in low dimensional problems.

\subsection{Forward problem}

As a benchmark problem, similarly to \cite{yang2018physics}, let us consider the following one-dimensional elliptic SDE
\begin{align}
\begin{split}
  -\frac{1}{10} \frac{\diff}{\diff x} [k(x,\omega) \frac{\diff}{\diff x} u(x,\omega) ] &= f(x,\omega), \quad x \in [-1,1],\\
  u(-1) = u(1) &= 0,
  \end{split}
  \label{eq.sde}
\end{align}

where on the domain boundary we  impose homogeneous Dirichlet boundary conditions on $u(x,  \omega)$.  The independent  stochastic processes $k(x,  \omega)$ and $f (x,  \omega)$ are given by

\begin{align}
\begin{split}
    k(x) &= \exp[\frac{1}{5} \sin(\frac{3 \pi}{2} (x+1)) + \hat{k}(x)], \\
    \hat{k}(x) &\sim \text{G\!P}(0, \frac{4}{25} \exp(-(x-x')^2)),\\
    f(x) &\sim \text{G\!P}(\frac{1}{2}, \frac{9}{400} \exp(-25(x-x')^2)),
      \end{split}
\end{align}

where $k(x,  \omega)$ is set to be strictly positive. In order to generate training data, we draw random sample paths of the stochastic processes $k(x,\omega)$ and $f(x,\omega)$, based on which  we use the finite difference  method to solve the corresponding deterministic differential equation to obtain realizations of the solution $u(x,\omega)$, as the reference. 

\begin{figure}
    \centering
    \includegraphics[width=16cm]{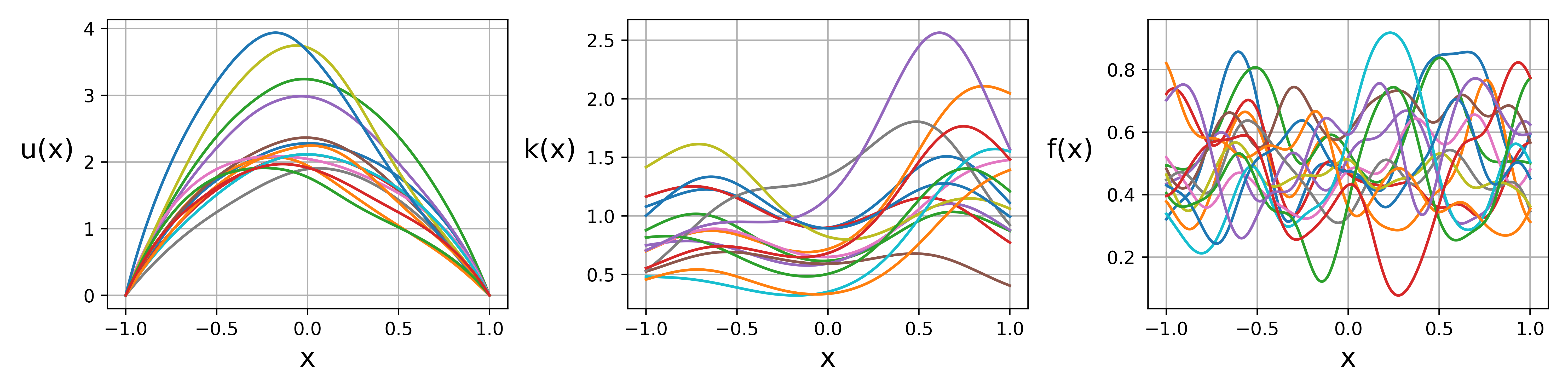}
    \caption{\footnotesize A few sample paths of $k(x,\omega)$, $u(x,\omega)$, and $f(x,\omega)$ used in the forward problem.}
    \label{fig.SDE_training_samples}
\end{figure}

In order to train the neural network models, we  use  17 $k$-sensors , 21 $f$-sensors, and 2 $u$-sensors, at locations uniformly distributed over the computational domain. The sample paths of $k(x,\omega)$, $u(x,\omega)$, and $f(x,\omega)$ are shown in Figure \ref{fig.SDE_training_samples}.

Our main quantity of interest in this problem is the mean and standard deviation of $u(x,\omega)$ on the validation coordinates, i.e., $\mu(x)=\mathbb{E}_w[u(x,\omega)]$ and $\sigma^2(x)= \mathbb{E}_w[(u(x,\omega)-\mu(x))^2]$. The number of validation points is 101. To study the influence of latent dimension, we fix the number of training snapshots to be 1000 and vary the latent dimension to be 2, 4, 10. Subsequently, we fix the latent dimension as 4 and vary the number of training snapshots to be 2000, 5000, and 10000 to study the influence of the number of training snapshots. During the training process, we keep the batch size to be 1000. One numerical solution of PI-WGAN using 10,000 snapshots with a four-dimensional latent space is obtained for comparison. 

\begin{figure}
\centering
\includegraphics[width=14cm]{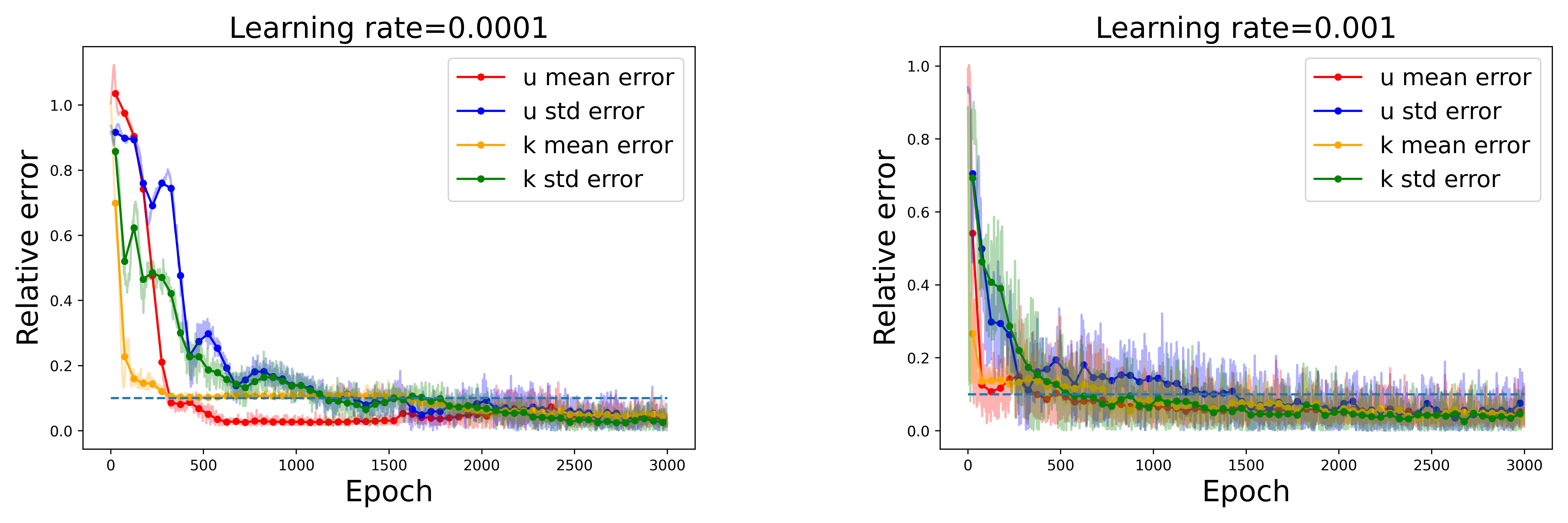}
\caption{\footnotesize Relative error of QoIs of PI-WGAN trained using different learning rate of 0.0001 (left) and 0.001 (right).}
\label{fig.GAN_of_diff_lr}
\end{figure}

Due to the instability of GAN training, the hyper-parameters of the training scheme should be carefully chosen. To ensure a fair comparison between PI-WGANs and our PI-VAE method, we first need to select appropriate hyper-parameters for the PI-WGAN training. Most of the parameters are chosen following \cite{yang2018physics}. Although the default value of the learning rate is 0.001, we observed that using a small learning rate (i.e. 0.0001) will result in better convergence of PI-WGAN, as shown in Figure \ref{fig.GAN_of_diff_lr}. The transparent curves show the exact error values, whereas the solid lines shown the  error averaged over 50 training epochs for better comparison. We call the accuracy satisfactory, when average error is less that 0.1.  In both  these experiments,  satisfactory accuracy leis achieved after approximately 1500 epochs, but the smaller learning rate has resulted in a more stable performance. Based on these two numerical experiments, we choose 0.0001 as the learning rate for PI-WGAN training.

\begin{figure}
\centering
\includegraphics[width=15cm]{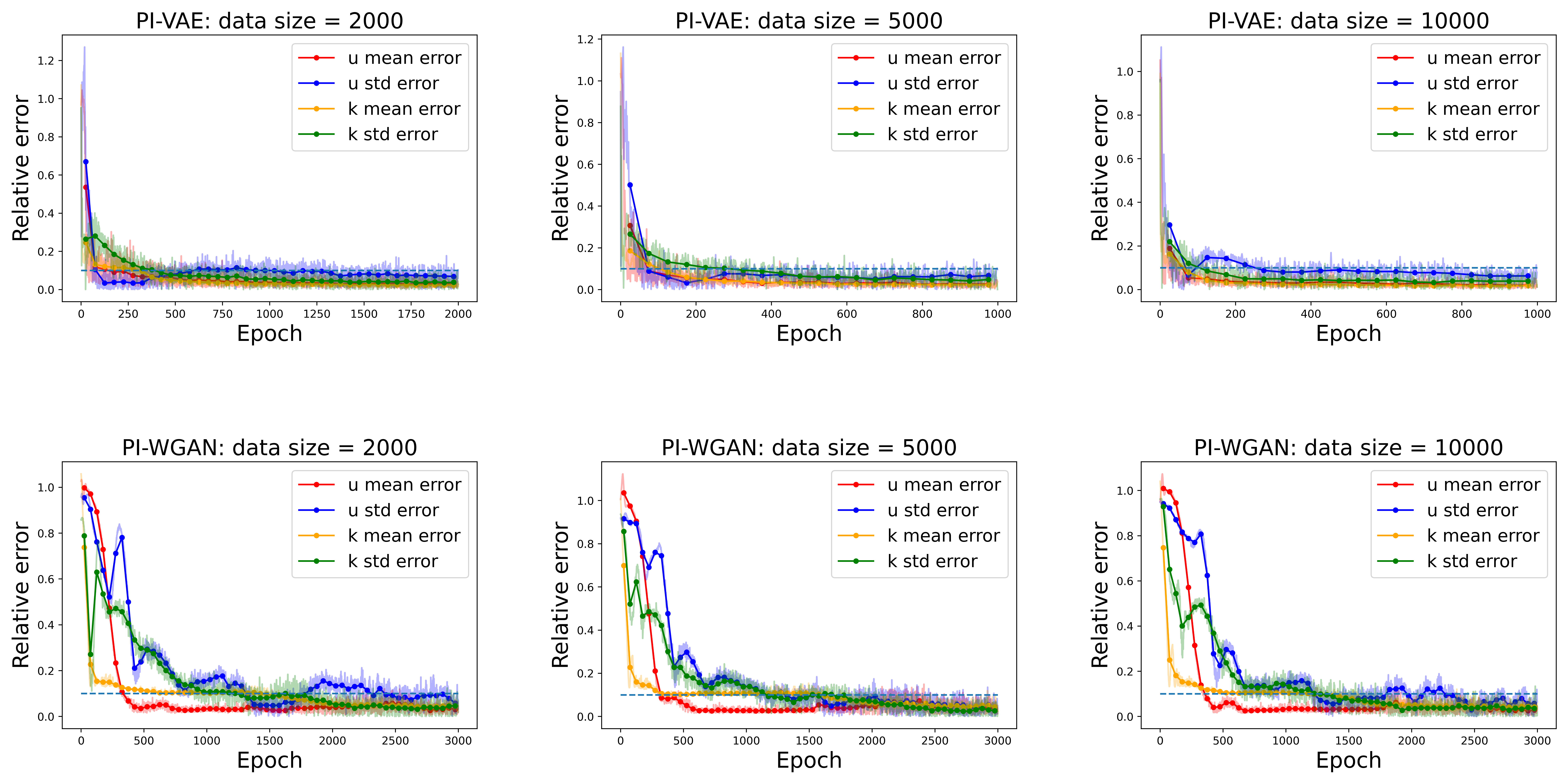}
\caption{\footnotesize Relative error of QoIs versus training epoch of PI-VAE (top) and PI-WGANs (bottom) using different training data size 2000 (left), 5000(mid), and 10000(right) are shown.}
\label{fig.forward_efficiency}
\end{figure}

Now let us compare the performance of PI-VAE against PI-WGAN  in terms of accuracy and efficiency. In Figure \ref{fig.forward_efficiency}, we present the error of QoI versus the training epoch. In our experiments, we  observed that only the training data size will affect the convergence and that different choices of dimension for the  latent space has limited effects. It can be seen, that even though a much larger learning rate is used for PI-VAE training, the model performance remains stable after achieving satisfactory accuracy. Also, the training of PI-VAE is more stable than the training of PI-WGAN since no over-fitting or abnormal "bounce" of the error  is detected in any of the numerical experiments. 

It should be noted that the definition of an epoch is different in the training of PI-WGAN and PI-VAE. Thus,  a more detailed efficiency comparison is presented in Table \ref{Time Estimate of converge time}. It should be noted that the computation cost of each training epoch is different in PI-VAE and PI-WGAN models. In the procedure of GAN training, we need to perform $n_d$ steps of optimizing discriminator parameters before a single update of generator parameters (with $n_de=5$ as the default value). This is because  we need to maintain the discriminator to be near-optimal during the whole training \cite{goodfellow2014generative}. However, VAE training is carried out similarly to the training of feed-forward neural networks. In other words, VAE  training involves only one forward propagation and one backward propagation over the encoder and decoder parts. In contrast,  GAN training involves  six  forward propagations over  the generator and discriminator, five  backward propagations over the discriminator, and one  backward propagation over the generator. 
 
Although the computation cost of backward will be higher, the computation cost of these two processes is at the same order of magnitude. Also, the architecture of the PI-VAE and PI-WGAN are similar, since we use the same neural network architecture between the encoder of PI-VAE and the discriminator of PI-WGAN, and also between the decoder of PI-VAE and the generator of PI-WGAN. As a result, the computational cost of one PI-WGAN training epoch is 4 to 5 times as large as the computational cost for one PI-VAE mini-batch training epoch.

Training is performed on an NVIDIA P100 GPU using the batch size of 1000. We need 0.62s for one training epoch of PI-WGAN while 0.13s for updating network parameters of PI-VAE using one mini-batch, which is consistent with our estimation. Based on the criterion mentioned before to evaluate the necessary training epoch of the model, we observe that at least 75\% of the computational cost can be saved if PI-VAE is used instead of PI-WGAN to solve this SDE.

\begin{table}
\caption{Comparison of converge time of different models(forward problem)}
\label{Time Estimate of converge time}
\centering
\begin{tabular}{Ic|c|c|cIc|c|c|cIcI}
\hline
model & \makecell{data \\ size} & \makecell{converge \\ epoch} & \makecell{time \\ (sec)} & model & \makecell{data \\ size} & \makecell{converge \\ epoch} & time & improvement\\
\hline
PI-VAE & 2000 & 1300 & 338 & PI-WGAN & 2000 & 2500 & 1550 & 4.58x \\
\hline
PI-VAE & 5000 & 400 & 260 & PI-WGAN & 5000 & 1800 & 1143 & 4.40x \\
\hline
PI-VAE & 10000 & 300 & 390 & PI-WGAN & 10000 & 2400 & 1560 & 4.00x \\
\hline
\end{tabular}
\end{table}

\begin{figure}
\centering
\includegraphics[width=15cm]{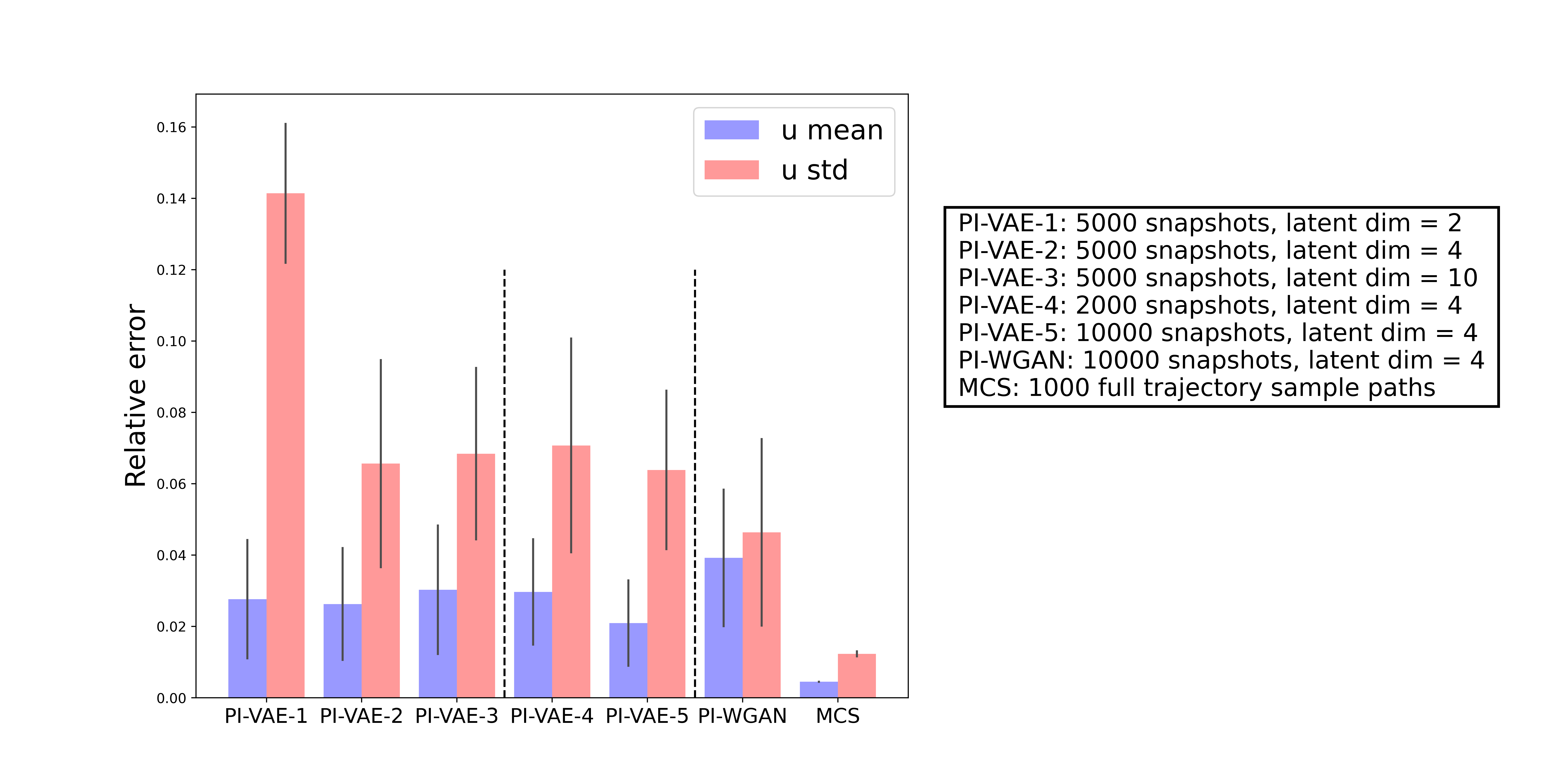}
\caption{\footnotesize Relative error of QoIs of different model for the forward problem is shown in this figure. Effect of number of snapshots and latent dimension on the accuracy of our method are shown in case 1 to case 5. Results of PI-WGANs and full trajectory sample paths in case 6 and case 7 are shown for comparison. }
\label{fig.forward_accuracy}
\end{figure}

For  accuracy evaluation, we use the approximation errors obtained in the last 100 training epochs as the criterion to evaluate the accuracy of the models. Figure \ref{fig.forward_accuracy} shows  the mean and standard deviation (std) of the QoIs calculate for different models. It can be seen that even though PI-WGAN has better accuracy in estimating the standard deviation, it offers lower accuracy in estimating the mean. Among the PI-VAE models, by varying the dimension of latent space in PI-VAE-1 to PI-VAE-3, we observe that the accuracy in the standard deviation estimates increase if we increase the dimension of the latent space, since more latent variables allow for more information of the high-dimensional input data to be encoded. This accuracy improvement is not significant beyond the dimensionality of 4. This is because higher dimension latent space is more difficult to sample, leading to a discrepancy between random samples and the prior distribution of latent space. It can also be seen from the figure, by comparing PI-VAE-2, PI-VAE-4, and PI-VAE-5, that larger number of training data also contributes to higher accuracy, although the effect is not as significant as that seen for different latent space dimensions. As shown in Figure \ref{fig.forward_accuracy}, the Monte Carlo Simulation (MCS) approach, which involves random sampling of the full trajectory, is the most accurate way to assess the statistics of QoIs, but it requires a large number of samples along the trajectory and is not practical in real-world applications.

\begin{figure}
\centering
\includegraphics[width=14cm]{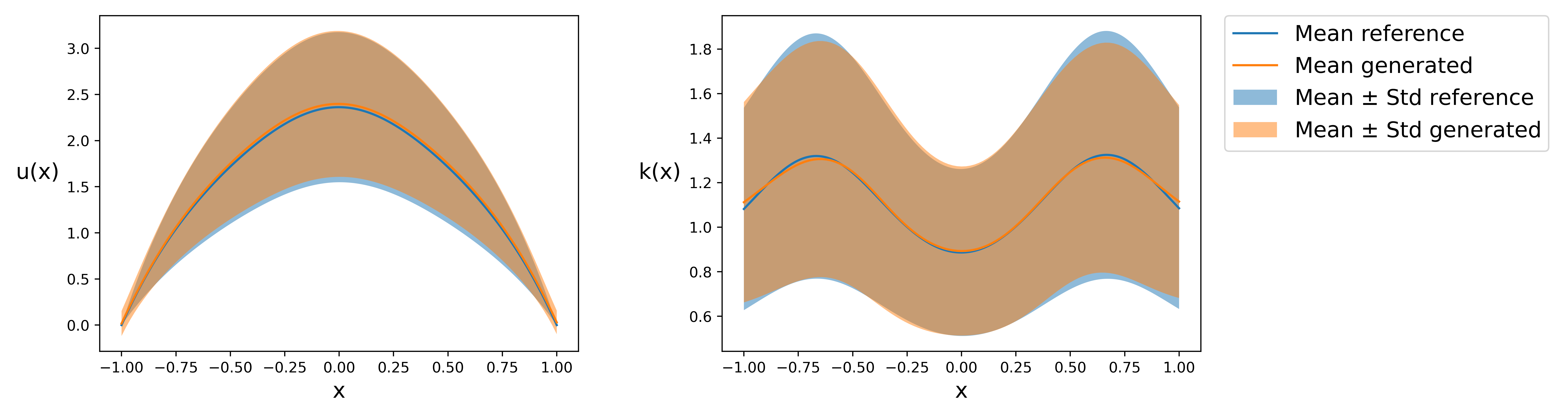}
\caption{\footnotesize Mean and variance estimate of QoIs using trained model in case 2 are shown. Blue line represent the reference mean and blue shade area represents the reference standard deviation. In contrast, orange line and orange shade area represent the mean and std of the reconstructed samples, respectively. }
\label{fig.forward_domain_plot}
\end{figure}

Next, we use the  trained PI-VAE-2 model to further evaluate the reproduced distribution for the process over the whole domain  and also at one particular location. Specifically, in Figure \ref{fig.forward_domain_plot} we show satisfactory accuracy in estimating the mean response and the 68\% confidence interval. The reference responses is calculated by obtaining another 10000 full trajectory sample paths. 

\begin{figure}
\centering
\includegraphics[width=7cm]{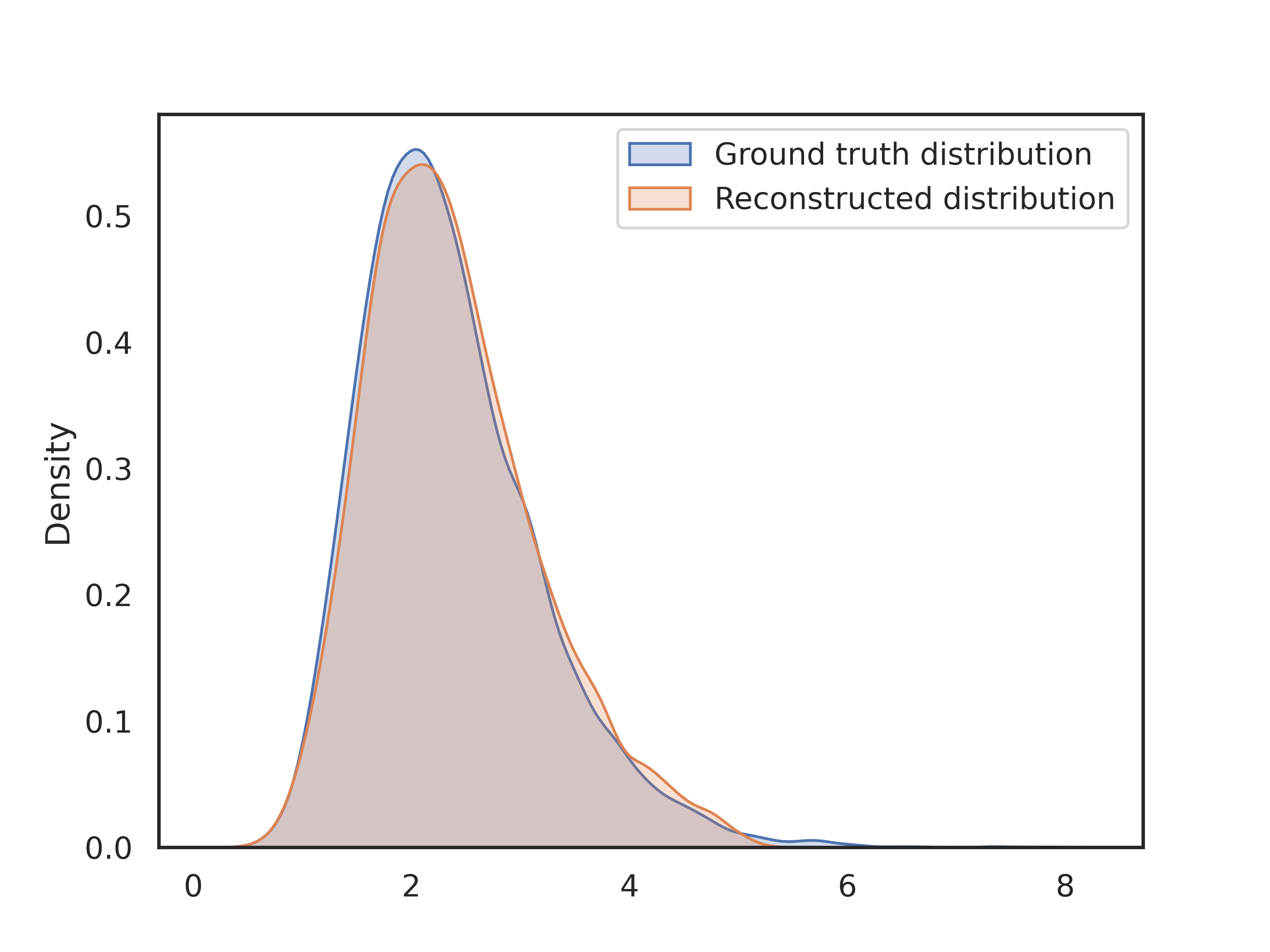}
\caption{\footnotesize Probability density distribution of the reconstructed samples and ground true samples of $u(x,\omega)$ at the coordinate of $x=0$ are shown.}
\label{fig.PDF_check}
\end{figure}

Also, the stochastic response $x=0$ is selected as another QoI. The reconstructed response is obtained by drawing 10,000 samples of latent variables and a feeding them to the trained decoder to generate  data points at $x=0$. It can be seen in Figure \ref{fig.PDF_check}, that the probability density functions (PDFs) of the reconstructed response is very close to that of the  ground truth.

\subsection{Inverse and mixed problems}

In this section, using the SDE formulation of Equation~\ref{eq.sde}, we  evaluate  the proposed PI-VAE method in solving a  range of problems, from forward problems to inverse problems and mixed problems in between. We consider $k(x,\omega)$ and $f(x,\omega)$ to be independent processes, and we study the following four problems each with a different sensor placement scenario. The setting of all scenarios are shown in Table \ref{table.sensor_placement} 


\begin{table}
\caption{Number of sensors used in different cases}
\label{table.sensor_placement}
\centering
\begin{tabular}{|c|c|c|c|c|c|}
\hline
{Case} & {Problem type} & {k-sensor} & \makecell{u-sensor \\ non-boundary} & \makecell{u-sensor \\ boundary} & {f-sensor} \\
\hline
{PI-VAE-1} & Inverse problem & 1 & 15 & 2 & 21 \\
\hline
{PI-VAE-2} & Mixed problem & 5 & 9 & 2 & 21  \\
\hline
 {PI-VAE-3} & Mixed problem& 11 & 3 & 2 & 21 \\
\hline
{PI-VAE-4} & Forward problem & 17 & 0 & 2 & 21 \\
\hline
\end{tabular}
\end{table}

\begin{figure}
    \centering
    \includegraphics[width=13cm]{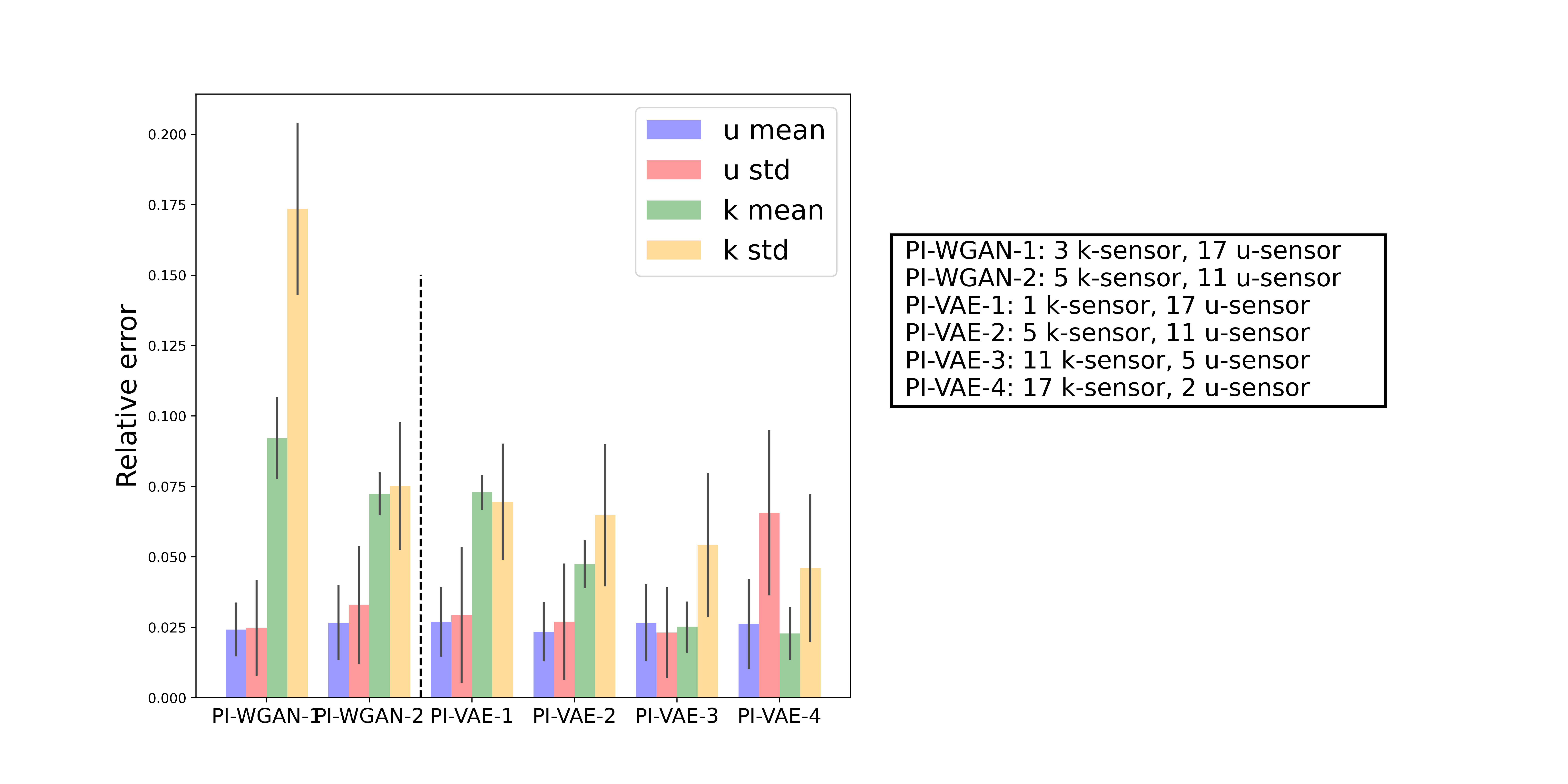}
    \caption{\footnotesize Relative error of QoIs of different model for the mixed problem and the inverse problem is shown in this figure. Results of PI-WGANs and in case 1 and case 2 are shown for comparison.}
    \label{fig.inverse_accuracy}
\end{figure}

We use $5,000$ snapshots as training set with a batch size of 1000. We select the 100 models from the last 100 epochs of the training to generate the reconstructed sample paths. The mean and standard deviation of $k(x,  \omega)$ and $u(x,  \omega)$ are still selected as QoIs and are shown in Figure \ref{fig.inverse_accuracy} for different trained models.    In the case of mixed problems, by comparing the results from PI-WGAN-2 and PI-VAE-2, that our proposed PI-VAE method is slightly more accurate. In solving inverse problem, by comparing the results of PI-WGAN-1 against PI-VAE-1, one can see that PI-VAE offers significantly higher accuracy levels at a much shorter training time.  It can also be seen  the errors in estimating $k(x,\omega)$ statistics will  decrease as the problem setup is changed from an inverse problem setting to a forward problem setting, by exploiting more information from $k(x,\omega)$. 

\begin{figure}
    \centering
    \includegraphics[width=9.5cm]{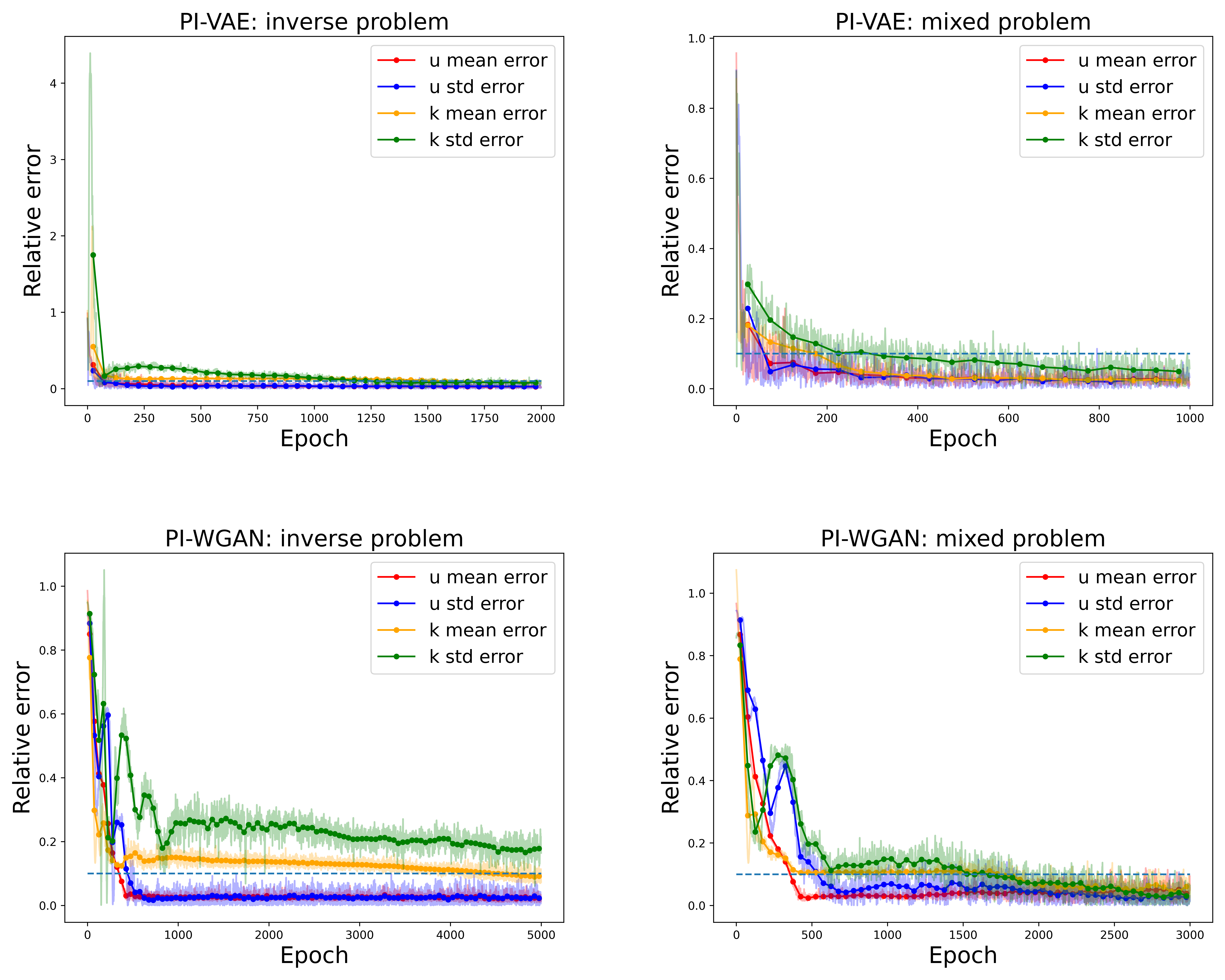}
    \caption{\footnotesize Relative error of QoIs versus training epoch of PI-VAE (top) and PI-WGANs (bottom) of different cases: inverse problem (left), mixed problem (right) are shown.}
    \label{fig.inverse_efficiency}
\end{figure}

Figure \ref{fig.inverse_efficiency} shows approximation errors versus training epoch for the inverse problem (scenario 1) and a mixed problem (scenario 2) settings. It can be seen that PI-VAE consistently converges much faster than PI-WGAN in inverse and mixed problems. In these experiments, we observed over 80\% of computational time saving when PI-VAE replaced PI-WGAN in solving these two cases.

\subsection{High-dimensional problems}

\begin{figure}
    \centering
    \begin{subfigure}[b]{0.8\textwidth}  
            \centering 
            \includegraphics[width=\textwidth]{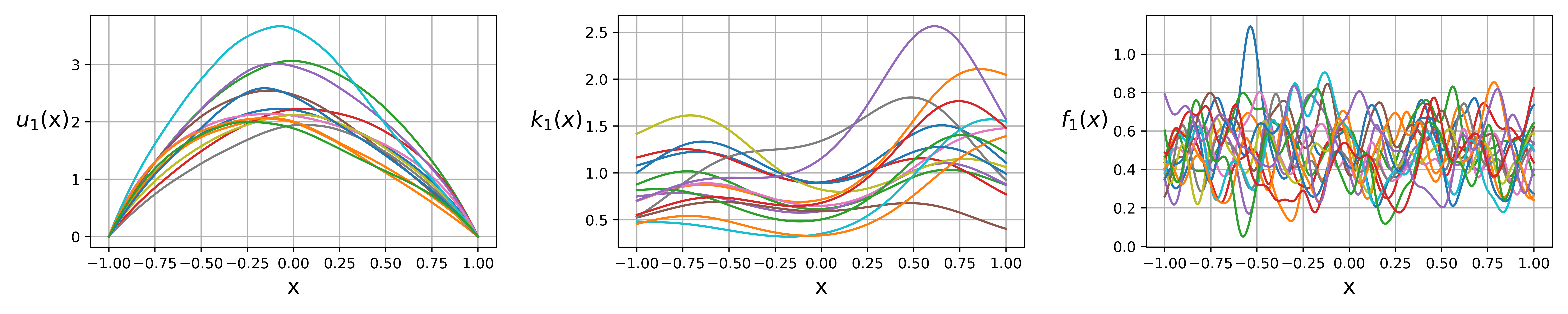}
            \caption{Training samples for the case with high dimensional $f$ }    
            \label{fig:training data of net1}
        \end{subfigure}
        \vskip\baselineskip
        \begin{subfigure}[b]{0.8\textwidth}   
            \centering 
            \includegraphics[width=\textwidth]{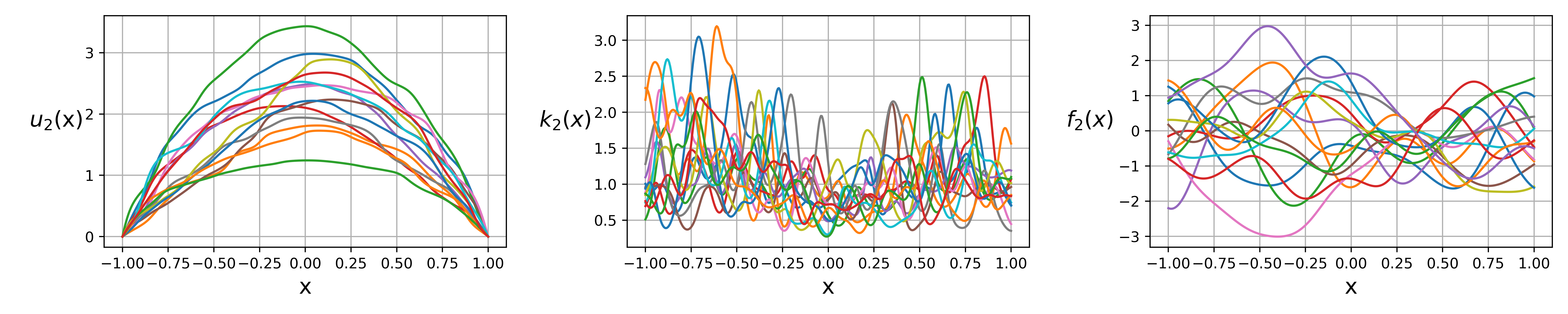}
            \caption{Training samples for the case with high dimensional $k$ }  
            \label{fig:training data of net2}
        \end{subfigure}

    \caption{\footnotesize Samples of response $u(x)$, $k(x)$ and $f(x)$ used in the training. The index 1 refers to the case where  $f$ has a high-dimensional representation and $k$ a low dimensional representation. Index 2 refers to the case where  the process $k$ is high-dimensional and $f$ is low-dimensional.}
    \label{fig.high_dimension_SDE_training_samples}
\end{figure}

In this section, we consider the same SDE formulation of Equation~\ref{eq.sde}, now with a \emph{high-dimensional} Gaussian process representation either for the coefficient term $k(x,\omega)$ or forcing term $f(x, \omega)$ and not both. This is to evaluate the ability of the PI-VAE method to solve the problems where there is an  imbalance between the dimensionality of the coefficient and forcing terms.  We do not analyze the case where both $k(x,\omega)$ and $f(x,\omega)$ are high-dimensional, as in those cases we  to place more sensors to measure these processes. The stochastic processes are formulated as:

\begin{align}
\begin{split}
    k(x) &= \exp[\frac{1}{5} \sin(\frac{3 \pi}{2} (x+1)) + \hat{k}(x)], \\
    \hat{k}(x) &\sim \text{G\!P}(0, \frac{4}{25} \exp( -\frac{1}{a^2}(x-x')^2)), \\
    f(x) &\sim \text{G\!P}(\frac{1}{2}, \frac{9}{400} \exp(-\frac{1}{b^2}(x-x')^2)),
    \end{split}
\end{align}

where $a$ and $b$ are the kernel length scales of the stochastic processes  $k(x, \omega)$ and $f(x, \omega)$, respectively, with the default values  $a=1$ and $b=1$. A process is transformed to a high-dimensional process by changing its length scale to the small value of 0.08. The sample paths of solutions $u_1(x, \omega)$ and $u_2(x, \omega)$, respectively corresponding to high-dimensional $f(x, \omega)$, $k(x, \omega)$, are shown  in Figure \ref{fig.high_dimension_SDE_training_samples} together with the sample paths of $f(x, \omega)$, $k(x, \omega)$. The number of sensors for a high-dimensional processes, whether $k(x, \omega )$ or $f(x, \omega)$, is set to be 51 to capture sufficient information. When a low dimensional process is concerned, the number of sensors remains the same as previous setting (i.e. 17 $k$-sensors, 21 $f$-sensors, and 2 $u$-sensors on the boundary).

\begin{figure}
    \centering
    \includegraphics[width=10.5cm]{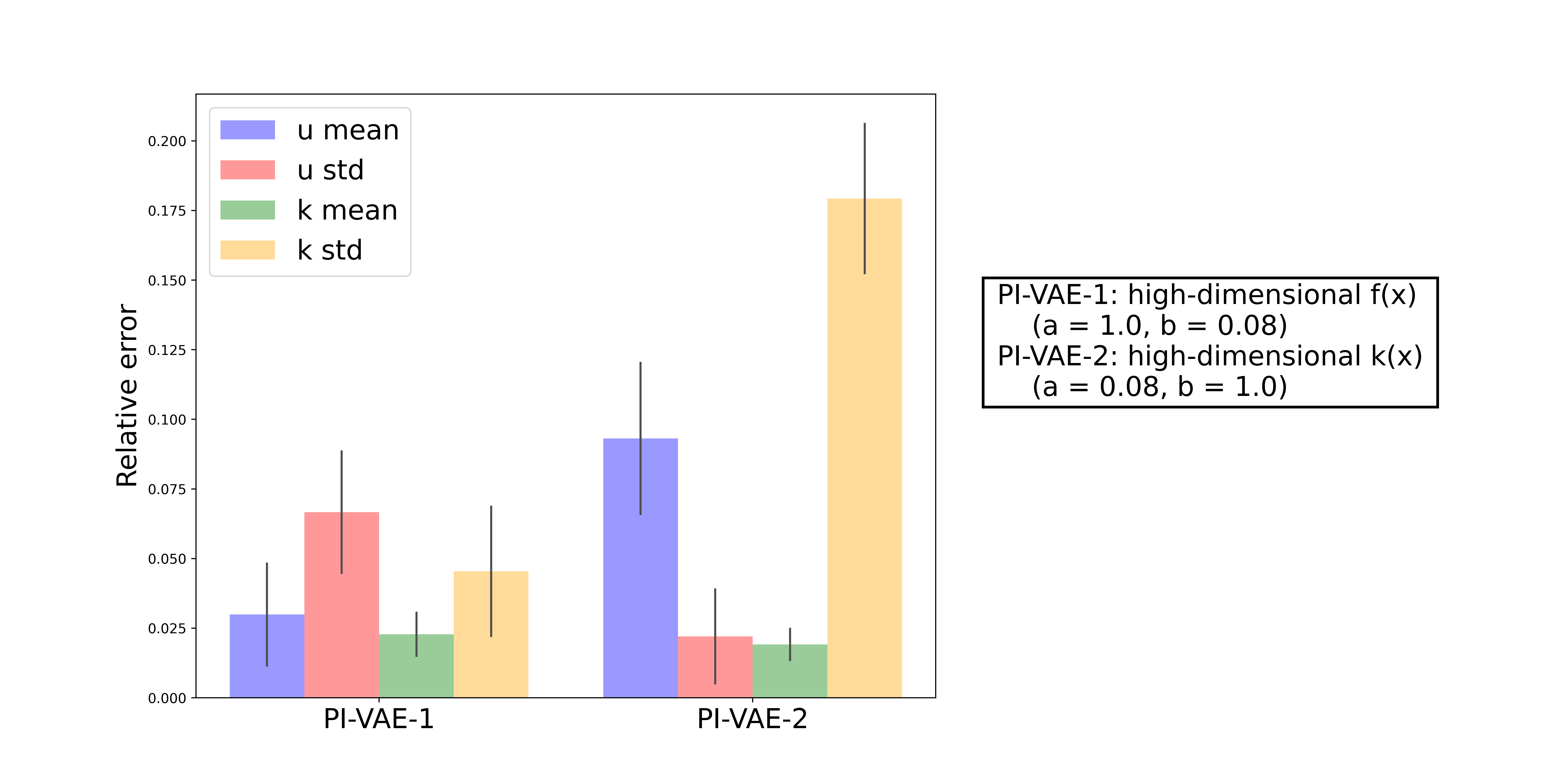}
    \caption{\footnotesize Relative error of QoIs of different model for the high-dimensional $k(x,\omega)$ problems (left) and high-dimensional $f(x,\omega)$ problems (right) is shown. }
    \label{fig.high_dimension_accuracy}
\end{figure}

We use $5,000$ snapshots as the training data, and we set the batch size to be 1,000. We train PI-VAE for 1,000 training epochs and train PI-WGAN for 5,000 training epochs, indicating that parameters of the decoder of PI-VAE and the generator of PI-WGAN go through the same number of  parameter updates. It should be noted that solving the problem with high-dimensional $k(x)$ is  more difficult than that with high-dimensional $f(x)$, since the derivative information of $k(x)$ is incorporated in the training loss. It can be seen in Figure \ref{fig.high_dimension_accuracy} that we can obtain much higher accuracy in the case of high-dimensional $f(x,\omega)$. Also, the results from the high-dimensional $f(x,\omega)$ case is even more accurate  compared to the results from low dimensional processes, mainly because more sensor information is obtained from the external forcing term. Except for the variance estimation of  $k(x,\omega)$, all QoIs satisfy the desired accuracy level.

\begin{figure}
    \centering

            \begin{subfigure}[b]{0.475\textwidth}
            \centering
            \includegraphics[width=\textwidth]{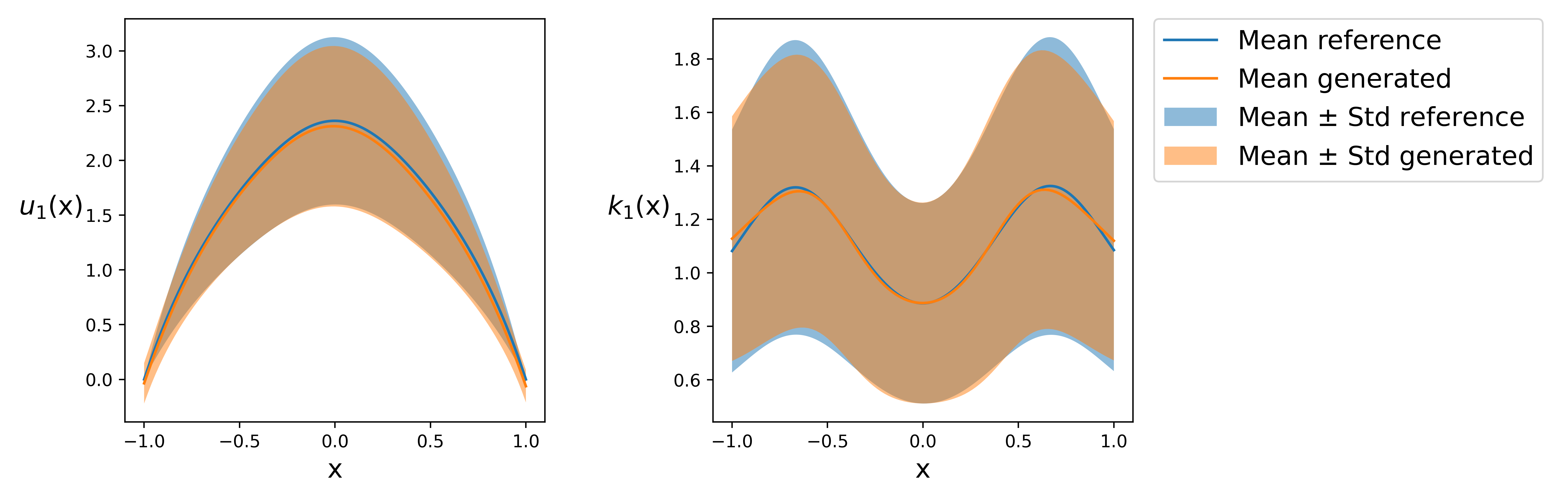}
            \caption{{PI-VAE: high dimensional $f$}}    
            \label{fig:mean and std of net1}
        \end{subfigure}
        \hfill
        \begin{subfigure}[b]{0.475\textwidth}  
            \centering 
            \includegraphics[width=\textwidth]{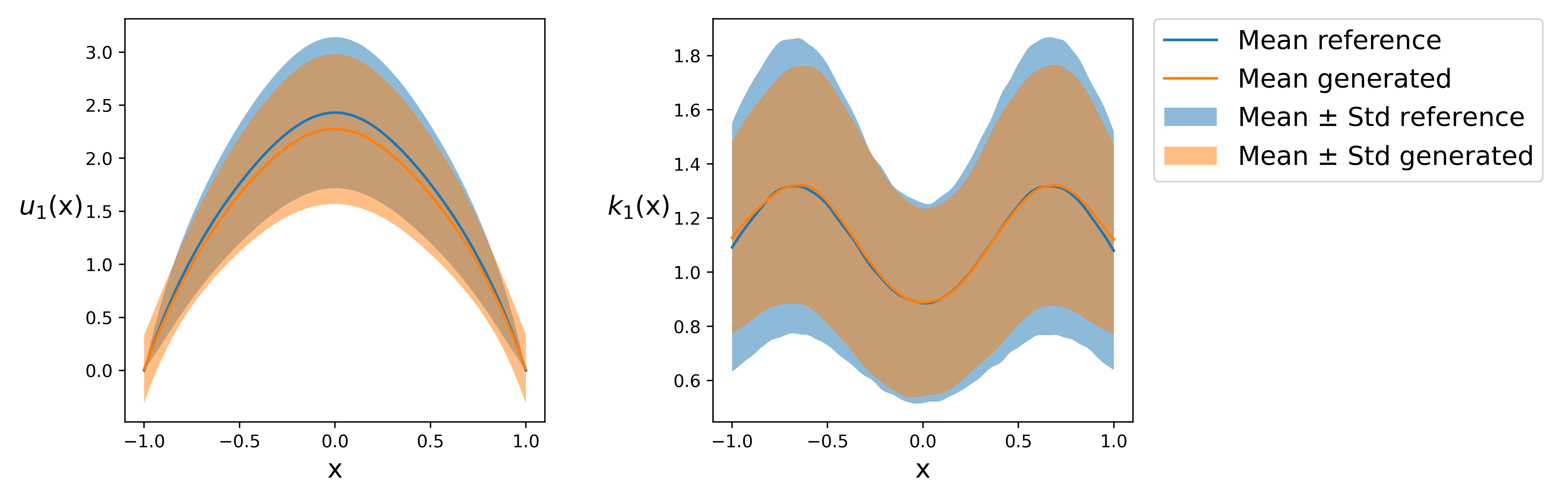}
            \caption{{PI-VAE: high dimensional $k$}}    
            \label{fig:mean and std of net2}
        \end{subfigure}
        \vskip\baselineskip
        \begin{subfigure}[b]{0.475\textwidth}   
            \centering 
            \includegraphics[width=\textwidth]{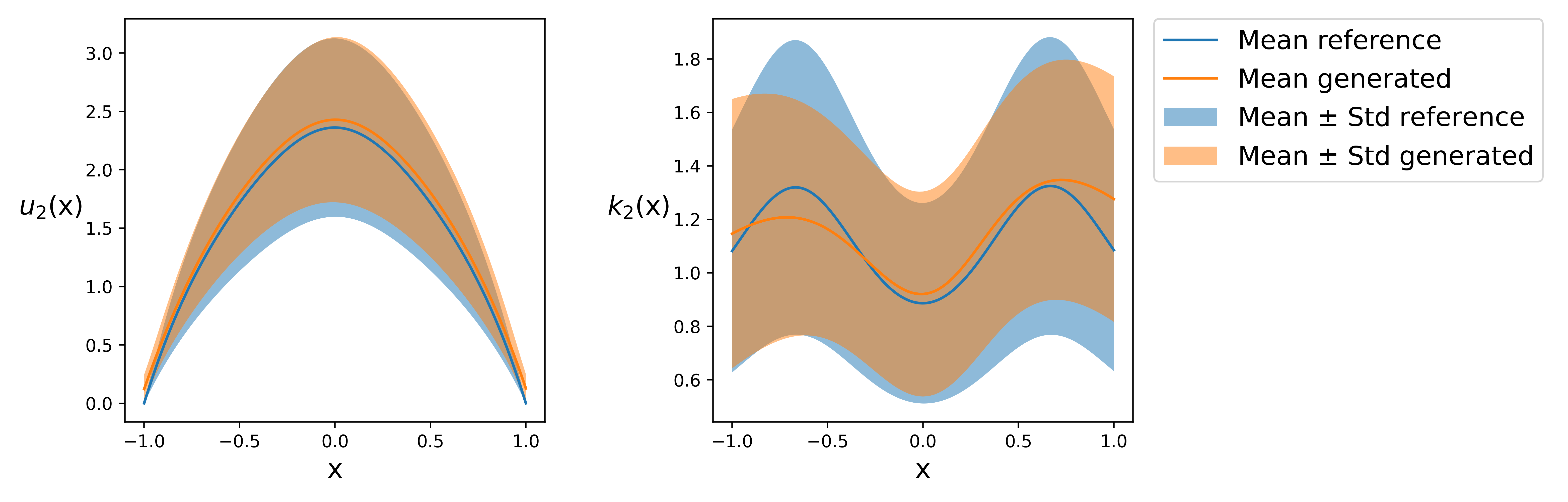}
            \caption{PI-WGAN: high dimensional $f$}  
            \label{fig:mean and std of net3}
        \end{subfigure}
        \hfill
        \begin{subfigure}[b]{0.475\textwidth}   
            \centering 
            \includegraphics[width=\textwidth]{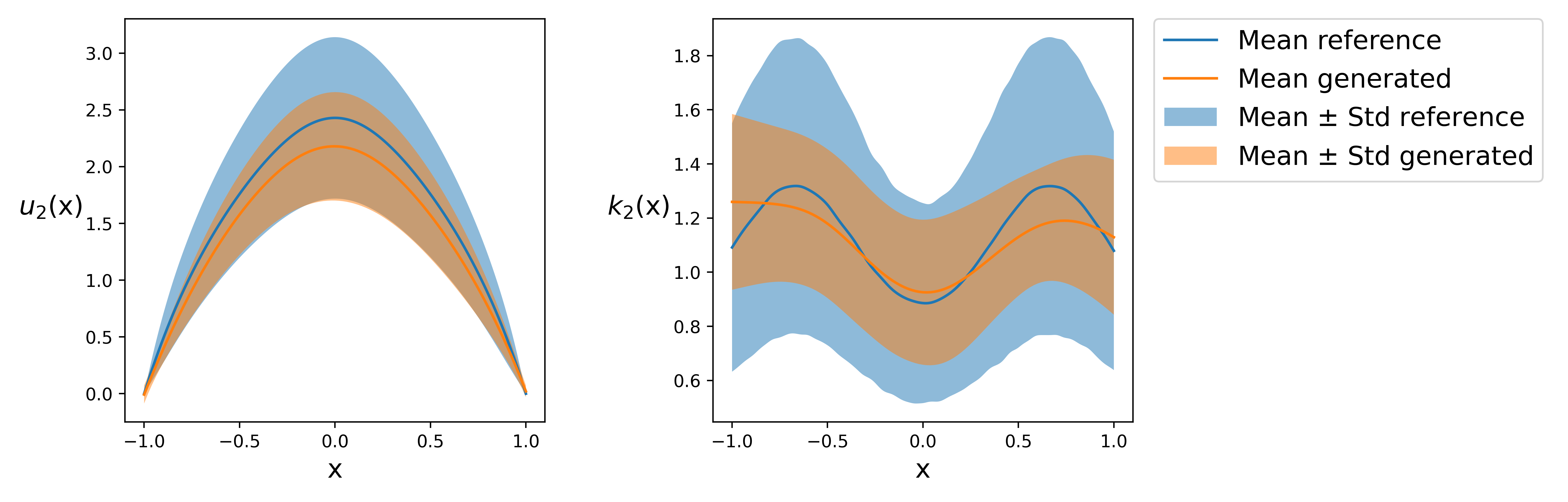}
            \caption{PI-WGAN: high dimensional $k$}  
            \label{fig:mean and std of net4}
        \end{subfigure}

    \caption{\footnotesize Mean and variance estimate of QoIs using trained model of PI-VAE (top) and PI-WGAN (bottom) for high-dimensional $k(x,\omega)$ problems (left) and high-dimensional $f(x,\omega)$ problems (right) are shown. Blue line represent the reference mean and blue shade area represents the reference standard deviation. In contrast, orange line and orange shade area represent the mean and std of the reconstructed samples, respectively.  }
    \label{fig.high_dimension_SDE_std_plot}
\end{figure}

In comparison with PI-WGAN, Figure \ref{fig.high_dimension_SDE_std_plot} shows that PI-VAE offers significantly better performance than PI-WGAN, especially  in the problem with high-dimensional $k(x,\omega)$. The disadvantage of PI-WGAN can be explained to be due to the gradient penalty \cite{gulrajani2017improved} in the loss function that constrains the gradient magnitude of the trained neural networks. In the case of high-dimensional $k(x,\omega)$, if we constrain the norm of the gradient, we cannot approximate $k(x,\omega)$ well enough even with sufficiently large number of $k$-sensors. In particular,  we observed that the estimation error for $k$ obtained from PI-VAE over the range of $x$ is at most 18\%, while PI-WGAN has estimation errors above 50\%.  

\section{Conclusion}\label{sec.conclusions}

In this paper, we focused on solving problems governed by PDEs where  the governing differential equation is known to us while information about the solution or the parameters of the differential equation are either unavailable, or  partially available from sparse measurement sensors.  We proposed PI-VAE; a new class of physics-informed Neural networks to efficiently solve these problems using inspirations from variational autoencoders. In comparison with PI-WGAN, which is another generative approach for solving similar problems, the presented  results in this paper show that our method can obtain satisfactory accuracy with less training time in most of the studied cases. Furthermore, the presented numerical results showed our method was more effective in forward, inverse and mixed problems, and also problems where random parameters of the governing physics are represented by a high-dimensional stochastic model. 

However,  limitations exist in our method. First of all, the accuracy of the loss function in our training depends on the size of the mini-batch. In particular, in high-dimensional differential equations where we need to differentiate the neural network many times, the GPU constraint will require us to use a small batch size in the training. Also, we now assume that all sensors will give us accurate information while, in practice, measurement noises exist and should be accounted for. Finally, even though our method achieved better accuracy levels compared to PI-WGAN in all the cases,  in a few of  numerical experiments presented in the paper, the performance levels can still be improved in further research extensions. Future research directions also include quantifying the the sensor uncertainty as well as the modeling uncertainty into the framework. 

\bibliographystyle{unsrt}  
\bibliography{references}  

\end{document}